\documentclass{article} % For LaTeX2e
\usepackage{PRIMEarxiv}

\usepackage{amsmath}
\usepackage{amsthm}
\usepackage{amsopn}
\usepackage{amsfonts}
\usepackage{amssymb}
\usepackage{mathrsfs}
\usepackage{bbm}
\usepackage{enumitem}
\usepackage{hyperref}
\usepackage{url}
\usepackage{graphicx}
\usepackage{subcaption}
\usepackage{algorithm}
\usepackage{algorithmic}
\usepackage{natbib}
\usepackage{booktabs}
\usepackage{cleveref}
\usepackage{todonotes}
\usepackage{pifont}

\usepackage{localmacros}

\newtheorem{theorem}{Theorem}

\newtheorem{remark}{Remark}

\title{Manifold Learning with Normalizing Flows: \\ 
Towards Regularity, Expressivity \\
and Iso-Riemannian Geometry}
\author{Willem Diepeveen \\
Department of Mathematics\\
University of California, Los Angeles\\
Los Angeles, CA 90095, USA \\
\texttt{wdiepeveen@math.ucla.edu} \\
\And
Deanna Needell \\
Department of Mathematics\\
University of California, Los Angeles \\
Los Angeles, CA 90095, USA \\
\texttt{deanna@math.ucla.edu}
}

\begin{document}

\maketitle

\begin{abstract}
    Modern machine learning increasingly leverages the insight that high-dimensional data often lie near low-dimensional, non-linear manifolds, an idea known as the manifold hypothesis. By explicitly modeling the geometric structure of data through learning Riemannian geometry algorithms can achieve improved performance and interpretability in tasks like clustering, dimensionality reduction, and interpolation. In particular, learned pullback geometry has recently undergone transformative developments that now make it scalable to learn and scalable to evaluate, which further opens the door for principled non-linear data analysis and interpretable machine learning. However, there are still steps to be taken when considering real-world multi-modal data. This work focuses on addressing distortions and modeling errors that can arise in the multi-modal setting and proposes to alleviate both challenges through isometrizing the learned Riemannian structure and balancing regularity and expressivity of the diffeomorphism parametrization. We showcase the effectiveness of the synergy of the proposed approaches in several numerical experiments with both synthetic and real data.
\end{abstract}

\blfootnote{Our code is available at \href{https://github.com/wdiepeveen/iso-Riemannian-geometry}{https://github.com/wdiepeveen/iso-Riemannian-geometry}.}

\section{Introduction}
\label{sec:intro}

\begin{figure}[h!]
    \centering
        \includegraphics[width=0.49\linewidth]{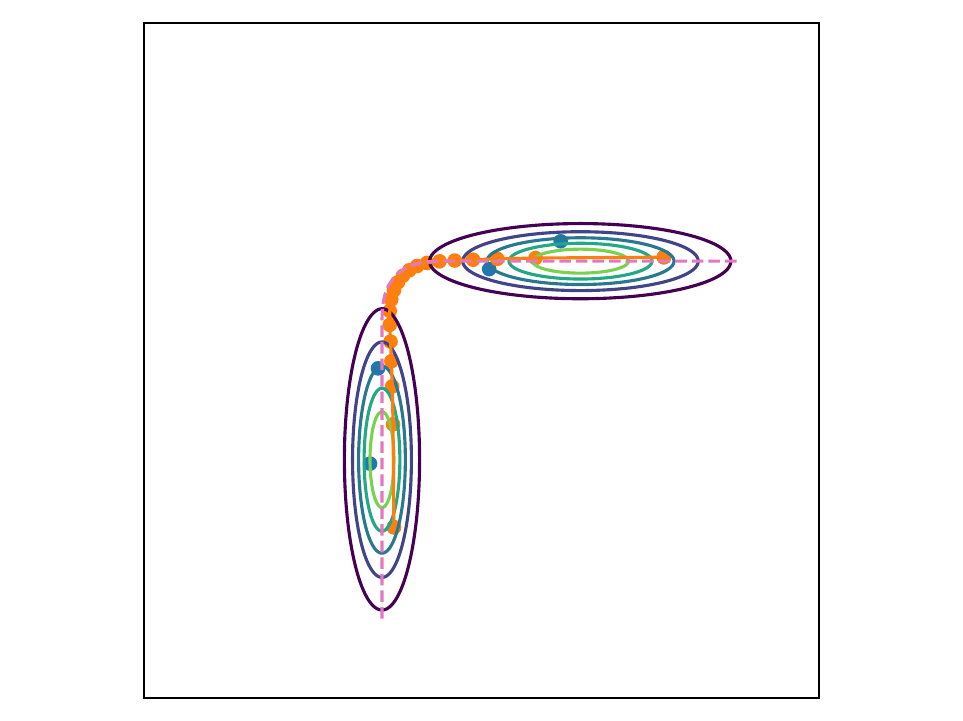}
        \caption{The $\Real^2$-valued data in blue, living in a bimodal normal distribution that is visualized by its level sets, reside closely to the manifold in pink (dashed). The geodesic in orange under a data-driven Riemannian structure (see \Cref{app:illustrative-pullback}) follows the manifold (and the regions of high likelihood) in the underlying data distribution in a natural fashion.}
        \label{fig:double-gaussian-data-analysis-results-toy-metric-geo}
\end{figure}

%% Geometry in machine learning
Modern machine learning methods increasingly rely on the assumption that high-dimensional data reside near a low-dimensional non-linear manifold (such as in Figure~\ref{fig:double-gaussian-data-analysis-results-toy-metric-geo}). This manifold hypothesis \cite{Fefferman2016} not only provides theoretical justification, but also significantly enhances algorithmic performance. For instance, using non-linear distances that reflect arc length along the manifold \cite{Little2022}, rather than Euclidean distances, improves clustering results \cite{Trillos2023}. Similarly, incorporating non-linearity into principal component analysis allows for better estimation of the intrinsic dimensionality of the data manifold \cite{Gilbert2025}, and explicitly imposing a Riemannian structure on the ambient space enables both of these benefits while facilitating interpolation and extrapolation that remain faithful to the data geometry \cite{Diepeveen2024b}. 

% [Modern machine learning methods for processing and analyzing data rely more and more on the assumption that data reside near a low-dimensional non-linear manifold. This manifold hypothesis \cite{Fefferman2016} and having implicit or explicit access to it, is not just a theoretical justification, but also really improves downstream algorithms. For example: having non-linear distances that capture the arc length distance over the manifold \cite{Little2022} (rather than the l2 distance) is more natural and in turn results in better clustering \cite{Trillos2023}; allowing for non-linearity in PCA, allows to capture the intrinsic dimension of the data manifold better \cite{Gilbert2025}; and more explicitly putting a Riemannian structure on ambient space allows for both of these, but also interpolation and extrapolation that are more faithful to the data \cite{Diepeveen2024b}. Whereas there are many ways to incorporate geometry into machine learning approaches \cite{Weber2025}, we will focus in this work on the latter example, i.e., using Riemannian geometry for machine learning with the distinction that the Riemannian structure is data-driven as well. This has been getting a lot of attention recently for the reason that it is a general language in which all of the above (and more) tasks can be done \cite{Gruffaz2025}.]
% \cite{Fefferman2016}
% \cite{Little2022}
% \cite{Gilbert2025}
% \cite{Weber2025}

% Riemannian geometry in machine learning
% \cite{Diepeveen2024b}
% \cite{Gruffaz2025}

While there are many ways to integrate geometric insights into machine learning approaches \cite{Weber2025}, Riemannian geometry, specifically in a data-driven setting, has gained significant attention recently due to its ability to provide a unified framework for addressing a wide range of tasks, including those mentioned above and beyond \cite{Gruffaz2025}. The aim of data-driven Riemannian geometry -- as we consider it in this work -- is to learn a Riemannian structure on the ambient space with a low-dimensional geodesic subspace that coincides with the data manifold, i.e., so that geodesics connecting data points always pass through the support of the data distribution (such as in Figure~\ref{fig:double-gaussian-data-analysis-results-toy-metric-geo}). 

Constructing such a Riemannian structure can be achieved in various ways. Early efforts that attempted to directly construct metric tensor fields \cite{Arvanitidis2016,Hauberg2012,Peltonen2004} were not very scalable on real (high-dimensional) data, both in terms of learning the geometry in the first place and evaluating manifold mappings subsequently. Later approaches introduced specialized structures to improve scalability for learning the geometry, either by making assumptions about the data \cite{Scarvelis2023} or about the Riemannian structure itself \cite{Sorrenson2024}. However, scalability of manifold mappings remained a challenge. The first step here was made through learning a pullback structure \cite{Diepeveen2024} derived from pairwise distances \cite{Fefferman2020}. Such Riemannian manifolds come with closed-form manifold mappings and stability guarantees when the pullback structure is a local $\ell^2$-isometry on the data support. However, in \cite{Diepeveen2024} the authors also faced a limitation in their proposed approach, specifically its lack of scalability when learning the geometry. Only recently, this gap between scalable training and scalable manifold mappings was bridged in \cite{Diepeveen2024a}, in which the authors demonstrated that a diffeomorphism obtained through adapted normalizing flow training \cite{Dinh2017} is an effective method for constructing pullback geometry\footnote{It should be noted, however, that \cite{Flouris2023} had previously observed that properly regularized normalizing flows could yield meaningful latent space interpolations.}. 

The key insight from \cite{Diepeveen2024a} for normalizing flow-based pullback geometry is that diffeomorphisms must balance regularity (for ensuring stability) and expressivity (for learning complex data manifolds), which in their analysis boils down to whether or not the diffeomorphisms are local $\ell^2$-isometries on the data support -- in line with earlier findings in \cite{Diepeveen2024}. This contrasts with normalizing flows as generative models \cite{Kobyzev2020}, which require non-volume-preserving diffeomorphisms to achieve universal distribution approximation \cite{teshima2020coupling}, inherently violating local isometry. While early architectures like additive flows \cite{Dinh2014} were actually volume-preserving, subsequent research shifted toward expressive non-volume-preserving designs, including affine couplings \cite{Dinh2017}, autoregressive flows \cite{Kingma2016}, and spline flows \cite{Durkan2019}, a trend sustained in recent advances \cite{Behrmann2019,Draxler2024,Kingma2018,Kolesnikov2024,Zhai2024}. Efforts to enhance volume-preserving flows \cite{Tomczak2017,Tomczak2016} have been overshadowed by the empirical dominance of non-volume-preserving approaches in density estimation tasks.

The tension between the type of flows needed for data-driven Riemannian geometry and the ones needed for generative modeling motivates a push for flow development that caters better to the needs of data-driven Riemannian geometry. In particular, this tension is stark in \cite{Diepeveen2024a}, where affine and spline flows are used for their expressivity, but achieve local isometry only under heavy regularization -- slowing training and still failing to guarantee local isometry in data-sparse regions, which can be a particularly problematic shortfall for downstream applications that value interpretability or fairness \cite{Mehrabi2021}. So, on one hand, this could be motivation to employ additive flows \cite{Dinh2014} instead. Yet, discarding advancements in normalizing flows is counterproductive, as parameter-prediction frameworks from affine flows could be repurposed for additive flows.

In this work, we set out to address two questions that are integral for the development of manifold learning with normalizing flows: (i) When we need to trade regularity for expressivity, can we construct systematic methods to retain balance by somehow ``isometrizing" the geometry while preserving the pullback manifold structure? (ii) Can recent developments -- including overshadowed regular linear architectures \cite{Tomczak2017,Tomczak2016} and expressive non-volume-preserving innovations \cite{Kingma2018,Kolesnikov2024,Zhai2024} -- enable diffeomorphisms to model complex manifolds without sacrificing regularity?

\subsection{Contributions}
As we will see in the following, the answers to both questions above is a resounding yes. In particular:

\paragraph{Isometrizing a Riemannian structure} For any Riemannian structure, we propose a systematic way to reparametrize geodesics such that they have constant $\ell^2$-speed. We show that our isometrization approach does not only solve any issues one encounters when interpolating under a data-driven Riemannian structure, but also gives rise to an iso-Riemannian geometry\footnote{i.e., isometrized versions of all manifold mappings} and straightforward generalization of Riemannian data analysis methods to the proposed iso-Riemannian framework. 

\paragraph{Regular yet expressive pullback geometry} Next, to learn pullback structures from multimodal data, we propose a combination of several advances from the normalizing flow literature that maintain regularity of the diffeomorphism without sacrificing too much expressivity. Our framework promotes choosing the smoothest way to transition between modes, while allowing for complex data geometries.

\paragraph{Regularity, expressivity and iso-Riemannian geometry} Finally, 
our numerical experiments with both synthetic and real data demonstrate that integrating all best practices -- iso-Riemannian geometry and regular yet expressive pullback geometry -- is the most effective approach moving forward.
% we argue through numerical experiments with both synthetic and real data that combining all best practices is the way forward.

% \todo[inline]{The contributions will be roughly that:
% - to show that any proposed architecture does what is should do, we will first consider, what can go wrong and how to correct for it. In particular, we show that lack of isometry will be an issue + we will show how we can post-process a Riemannian structure to be more isometric in basic data analysis tasks
% - Subsequently, we will propose architectures and argue through numerical experiments that for our proposed architectures, the naive case gives similar results to the most corrected-for case.}

% - we will see that constant speed is enough (so local isometry is a bit too strong an assumption)
\subsection{Outline}
This article is structured as follows. Section~\ref{sec:notation} covers basic notation from diﬀerential, Riemannian and pullback geometry. In Section~\ref{sec:iso-riemannian-data-processing} we provide the reader with intuition as to why lack of isometry is problematic in Riemannian data processing, set up iso-Riemannian geometry and argue how it can be used in downstream tasks such as low-rank approximation.
Section~\ref{sec:regular-nfs} explores why diffeomorphisms without much regularity are prone to learn incorrect pullback geometry and focuses on leveraging ideas from the normalizing flow literature to alleviate such problems.
In Section~\ref{sec:numerics} we combine best practices and explore overall performance improvement through several numerical experiments. Finally, we summarize our findings in Section~\ref{sec:conclusions}.

% - sec 2 notation
% - sec 3 getting intuition in the problem with not having regularity and isometry and why it is not as straightforward to fix things
% - sec 5 regular parametrizations that are more likely to hold up in data sparse regions
% - sec 5 see how we can get isometry again, but also discuss limitations -- we will still need good geodesics --
% - sec 6 numerics
\section{Notation}
\label{sec:notation}

Here we present some basic notations from differential and Riemannian geometry, see \cite{boothby2003introduction,carmo1992riemannian,lee2013smooth,sakai1996riemannian} for details.

\paragraph{Smooth manifolds and tangent spaces} 
Let $\manifold$ be a smooth manifold. We write $C^\infty(\manifold)$ for the space of smooth functions over $\manifold$. The \emph{tangent space} at $\mPoint \in \manifold$, which is defined as the space of all \emph{derivations} at $\mPoint$, is denoted by $\tangent_\mPoint \manifold$ and for \emph{tangent vectors} we write $\tangentVector_\mPoint \in \tangent_\mPoint \manifold$. For the \emph{tangent bundle} we write $\tangent\manifold$ and smooth vector fields, which are defined as \emph{smooth sections} of the tangent bundle, are written as $\vectorfield(\manifold) \subset \tangent\manifold$.

\paragraph{Riemannian manifolds} 
A smooth manifold $\manifold$ becomes a \emph{Riemannian manifold} if it is equipped with a smoothly varying \emph{metric tensor field} $(\,\cdot\,, \,\cdot\,) \colon \vectorfield(\manifold) \times \vectorfield(\manifold) \to C^\infty(\manifold)$. This tensor field induces a \emph{(Riemannian) metric} $\distance_{\manifold} \colon \manifold\times\manifold\to\Real$. The metric tensor can also be used to construct a unique affine connection, the \emph{Levi-Civita connection}, that is denoted by $\nabla_{(\,\cdot\,)}(\,\cdot\,) : \vectorfield(\manifold) \times \vectorfield(\manifold) \to \vectorfield(\manifold)$. 
This connection is in turn the cornerstone of a myriad of manifold mappings.

One is the notion of a \emph{geodesic}, which for two points $\mPoint,\mPointB \in \manifold$ is defined as a curve $\geodesic_{\mPoint,\mPointB} \colon [0,1] \to \manifold$ with minimal length that connects $\mPoint$ with $\mPointB$. Another closely related notion to geodesics is the curve $t \mapsto \geodesic_{\mPoint,\tangentVector_\mPoint}(t)$  for a geodesic starting from $\mPoint\in\manifold$ with velocity $\dot{\geodesic}_{\mPoint,\tangentVector_\mPoint} (0) = \tangentVector_\mPoint \in \tangent_\mPoint\manifold$. This can be used to define the \emph{exponential map} $\exp_\mPoint \colon \mathcal{D}_\mPoint \to \manifold$ as \(\exp_\mPoint(\tangentVector_\mPoint) := \geodesic_{\mPoint,\tangentVector_\mPoint}(1),\) where \(\mathcal{D}_\mPoint \subset \tangent_\mPoint \manifold\) is the set on which \(\geodesic_{\mPoint,\tangentVector_\mPoint}(1)\) is defined. The manifold $\manifold$ is said to be \emph{(geodesically) complete} whenever $\mathcal{D}_{\mPoint}=\tangent_{\mPoint} \manifold$ for all $\mPoint \in \manifold$. Furthermore, the \emph{logarithmic map} $\log_\mPoint \colon \exp_\mPoint(\mathcal{D}'_\mPoint ) \to \mathcal{D}'_\mPoint$ is defined as the inverse of $\exp_\mPoint$, so it is well-defined on  $\mathcal{D}'_{\mPoint} \subset \mathcal{D}_{\mPoint}$ where $\exp_\mPoint$ is a diffeomorphism.

\paragraph{Pullback manifolds} 
Finally, if $(\manifold, (\cdot,\cdot))$ is a $\dimInd$-dimensional Riemannian manifold, $\manifoldB$ is a $\dimInd$-dimensional smooth manifold and $\diffeo:\manifoldB \to \manifold$ is a diffeomorphism, the \emph{pullback metric}
\begin{equation}
    (\tangentVector, \tangentVectorB)_\mPoint^\diffeo := (D_{\mPoint}\diffeo[\tangentVector_{\mPoint}], D_{\mPoint}\diffeo[\tangentVectorB_{\mPoint}])_{\diffeo(\mPoint)}, \quad \mPoint \in \manifold, \tangentVector, \tangentVectorB \in \vectorfield(\manifold)
    \label{eq:pull-back-metric}
\end{equation}
where $D_{\mPoint}\diffeo: \tangent_\mPoint \manifoldB \to \tangent_{\diffeo(\mPoint)} \manifold$ denotes the differential of $\diffeo$,
defines a Riemannian structure on $\manifoldB$, which we denote by $(\manifoldB, (\cdot,\cdot)^\diffeo)$. 
Pullback mappings are denoted similarly to (\ref{eq:pull-back-metric}) with the diffeomorphism $\diffeo$ as a superscript, i.e., we write $\distance^\diffeo_{\manifoldB}(\mPoint, \mPointB)$, $\geodesic^\diffeo_{\mPoint, \mPointB}$, $\exp^\diffeo_\mPoint (\tangentVector_\mPoint)$ and $\log^\diffeo_{\mPoint} \mPointB$ for $\mPoint,\mPointB \in \manifoldB$ and $\tangentVector_\mPoint \in \tangent_\mPoint \manifoldB$. Pullback metrics literally pull back all geometric information from the Riemannian structure on $\manifold$. 
In particular, closed-form manifold mappings on $(\manifold, (\cdot,\cdot))$ yield under mild assumptions closed-form manifold mappings on $(\manifoldB, (\cdot,\cdot)^\diffeo)$.

% \paragraph{Data analysis under Euclidean pullback.}

\section{Isometrized Riemannian data processing of real-valued data}
\label{sec:iso-riemannian-data-processing}

% \subsection{Why we need locally $\ell^2$-isometric Riemannian geometry.}
% \label{sec:naive-data-analysis}

We start with addressing the first question at the end of Section~\ref{sec:intro}: 
\begin{center}
    \emph{When we need to give in regularity for expressivity, can we construct systematic methods to retain balance by somehow ``isometrizing" the geometry while preserving the pullback manifold structure?}
\end{center}
However, before going into the details of our proposed way of isometrizing a general Riemannian structure, we aim to give the reader an understanding as to why lacking local $\ell^2$-isometry on the data support -- or more generally constant geodesic speed in an $\ell^2$-sense -- can potentially be an issue in the setting of multimodal data distributions. To illustrate this, we will consider problems arising in pullback geodesics and pullback non-linear low-rank approximation at the Riemannian barycentre, where the latter is used to construct an approximate data manifold and to retrieve its intrinsic dimension. Since these two tasks are the building blocks of Riemannian data analysis, having issues here is potentially detrimental for downstream applications.

\begin{figure}[h!]
    \centering
    \begin{subfigure}[b]{0.49\linewidth}
        \centering
        \includegraphics[width=\linewidth]{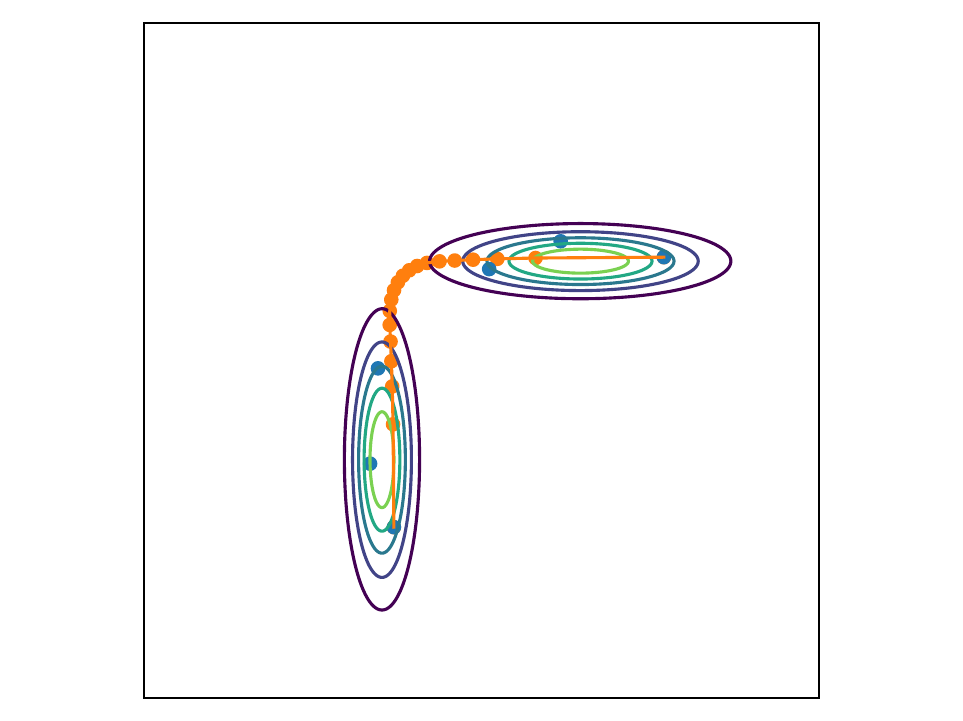}
        \caption{Modeled pullback geodesic}
        \label{fig:double-gaussian-data-analysis-results-affine-unbend-metric-geo}
    \end{subfigure}
    \hfill
    \begin{subfigure}[b]{0.49\linewidth}
        \centering
        \includegraphics[width=\linewidth]{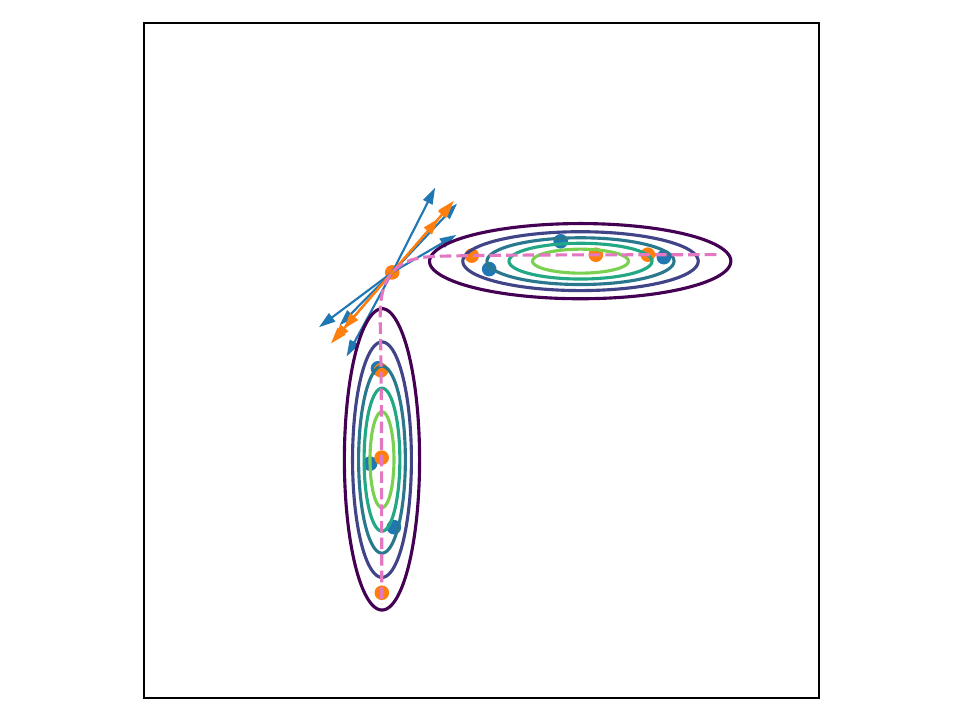}
        \caption{Modeled pullback rank-1 approximation}
        \label{fig:double-gaussian-data-analysis-results-affine-unbend-metric-pga}
    \end{subfigure}
    \caption{Both geodesic interpolation and non-linear rank-1 approximation under a pullback structure on $\Real^2$ used in Figure~\ref{fig:double-gaussian-data-analysis-results-toy-metric-geo} suffer from distortions that are detrimental for interpolate interpretability and rank-1 reconstruction performance.}
    \label{fig:double-gaussian-data-analysis-results-affine-unbend-metric}
\end{figure}

We will consider the Euclidean pullback structure used in Figure~\ref{fig:double-gaussian-data-analysis-results-toy-metric-geo} (see \Cref{app:illustrative-pullback} for details) and validate the pullback geometry on six validation data points in blue from the same Figure.
For Euclidean pullback, i.e., from a diffeomorphism $\diffeo:\Real^\dimInd\to \Real^\dimInd$, we have \cite[Prop~2.1]{Diepeveen2024}
\begin{align}
    &\geodesic^{\diffeo}_{\Vector, \VectorB}(t) = \diffeo^{-1}((1 - t)\diffeo(\Vector) + t \diffeo(\VectorB)),
    \label{eq:thm-geodesic-remetrized}\\
    &\exp^{\diffeo}_\Vector (\tangentVector_\Vector) = \diffeo^{-1}(\diffeo(\Vector) + D_{\Vector} \diffeo[\tangentVector_\Vector]),
    \label{eq:thm-exp-remetrized}\\
    &\log^{\diffeo}_{\Vector} \VectorB = D_{\diffeo(\Vector)}\diffeo^{-1}[\diffeo(\VectorB) - \diffeo(\Vector)],
    \label{eq:thm-log-remetrized}
\end{align}
where $\Vector, \VectorB\in \Real^\dimInd$ and $\tangentVector_\Vector \in \tangent_\Vector \Real^\dimInd \cong \Real^\dimInd$, and have \cite[Prop~3.7]{Diepeveen2024}
\begin{equation}
    \operatorname{argmin}_{\Vector\in \Real^\dimInd} \sum_{\sumIndA=1}^\dataPointNum \distance^\diffeo_{\Real^\dimInd}(\Vector, \Vector^\sumIndA)^2 = \diffeo^{-1} (\frac{1}{\dataPointNum} \sum_{\sumIndA=1}^\dataPointNum \diffeo(\Vector^\sumIndA)),
\end{equation}
for the Riemannian barycentre, where $\Vector^1, \ldots, \Vector^\dataPointNum\in \Real^\dimInd$. We highlight that this pullback structure is not locally $\ell^2$-isometric on the data support. The results shown in Figure~\ref{fig:double-gaussian-data-analysis-results-affine-unbend-metric} indicate that several things are going wrong: 

\paragraph{Distorted geodesics}
Because geodesics do not have constant speed in the $\ell^2$-sense\footnote{normally implied by local $\ell^2$-isometry on the data support}, the Riemannian structure underlying Figure~\ref{fig:double-gaussian-data-analysis-results-affine-unbend-metric-geo} creates the illusion that the data in the low-density region is more important as the geodesic spends a most of time here (visualized as the higher density of orange dots\footnote{representing time-equidistant points along the geodesic}). So when interpolating, one would see mostly data that is very rare to see. This is detrimental for interpretability when one tries to get intuition for what data points are ``typically in between'' the two end points. Even though this Riemannian structure is modeled, such a scenario or the converse where the speed is much higher in the low-density region is bound to happen as there are no data points to regularize for local isometry in regions without data.

\paragraph{Distorted dimension reduction}
This distortion is also detrimental when trying to do non-linear low-rank approximation. In particular, in Figure~\ref{fig:double-gaussian-data-analysis-results-affine-unbend-metric-pga} we see that the $\ell^2$-distances between tangent vectors in the tangent space do not correspond to (or carry any information for) the actual $\ell^2$-distance between the data points. So despite the fact that the Riemannian structure is able to learn a reasonable data manifold approximation, any error due to projection onto a one dimensional subspace of the tangent space is amplified for the reconstructions.

\begin{remark}
\label{rem:l2-pga-instead-of-pga}
    In the above we used non-linear low-rank approximation in the sense of Algorithm~\ref{alg:l2-tangent-space-SVD}, i.e., through low-rank approximation in a tangent space under the $\ell^2$-metric. The approach used is slightly different than the standard setting in \cite{fletcher2004principal}, where the inner product from the Riemannian structure is used rather than the $\ell^2$-inner product. The choice of Algorithm~\ref{alg:l2-tangent-space-SVD} is further justified later in this section.
\end{remark}

\begin{algorithm}[h!]
\caption{$\ell^2$-tangent space rank-$r$ approximation on $(\Real^\dimInd, (\cdot, \cdot))$}
\label{alg:l2-tangent-space-SVD}
\begin{algorithmic}
\STATE{\textit{Initialisation}: $r\in \Natural$, $\Matrix :=[\Vector^1, \Vector^2, \ldots, \Vector^\dataPointNum]\in \Real^{\dimInd \times \dataPointNum}$, $\mPoint\in\Real^{\dimInd}$}
\STATE Compute $\log_\mPoint(\Matrix) := [\log_\mPoint(\Vector^1), \log_\mPoint(\Vector^2), \ldots, \log_\mPoint(\Vector^\dataPointNum)]\in (\tangent_\mPoint \Real^\dimInd)^\dataPointNum \cong\Real^{\dimInd \times \dataPointNum}$
\STATE Compute the SVD $\log_\mPoint(\Matrix) = \mathbf{U}\Sigma \mathbf{V}^\top$
\STATE Compute rank-$r$ approximation $\tangentVector_\mPoint:= [\tangentVectorComp_\mPoint^1, \tangentVectorComp_\mPoint^2, \ldots, \tangentVectorComp_\mPoint^\dataPointNum] := \mathbf{U}_r\Sigma_r \mathbf{V}_r^\top$ from the top $r$ singular values and vectors
\STATE Compute rank-$r$ approximation $\tilde{\Matrix} := [\exp_{\mPoint}(\tangentVectorComp_\mPoint^1), \exp_{\mPoint}(\tangentVectorComp_\mPoint^2), \ldots, \exp_{\mPoint}(\tangentVectorComp_\mPoint^\dataPointNum)]\in \Real^{\dimInd \times \dataPointNum}$
\RETURN $\tilde{\Matrix}$
\end{algorithmic}
\end{algorithm}

\begin{figure}[h!]
    \centering
    \begin{subfigure}[b]{0.49\linewidth}
        \centering
        \includegraphics[width=\linewidth]{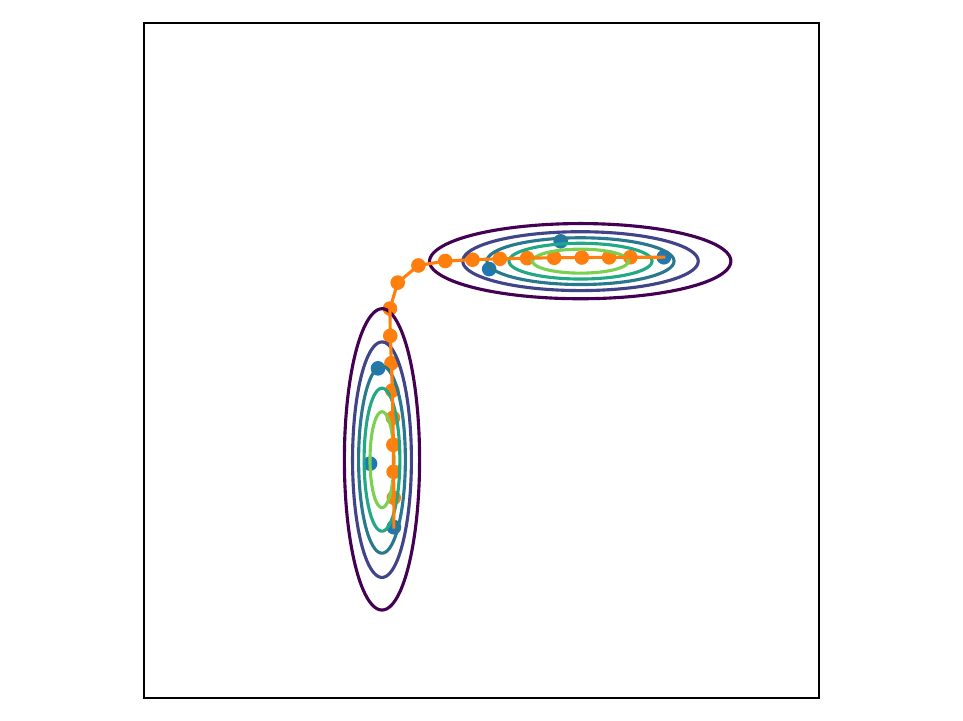}
        \caption{Isometrized pullback geodesic}
        \label{fig:double-gaussian-data-analysis-results-affine-unbend-metric-geo-iso}
    \end{subfigure}
    \hfill
    \begin{subfigure}[b]{0.49\linewidth}
        \centering
        \includegraphics[width=\linewidth]{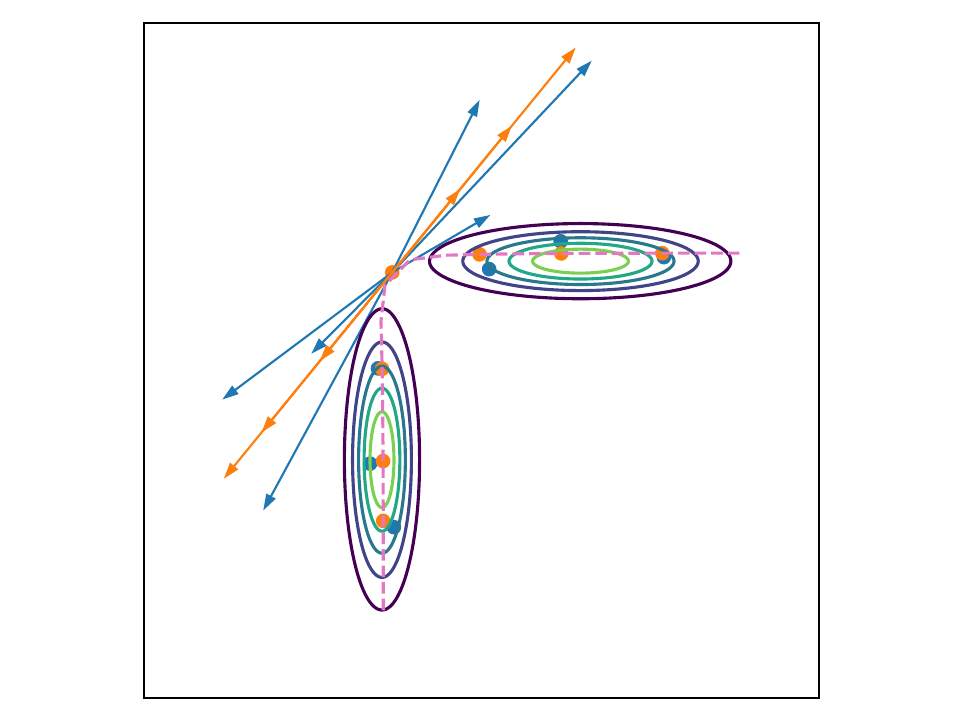}
        \caption{Isometrized pullback rank-1 approximation}
        \label{fig:double-gaussian-data-analysis-results-affine-unbend-metric-pga-iso}
    \end{subfigure}
    \caption{Neither isometrized geodesic interpolation nor isometrized non-linear rank-1 approximation under a pullback structure on $\Real^2$ suffer from distortions that the initial pullback structure suffered from (see Figure~\ref{fig:double-gaussian-data-analysis-results-affine-unbend-metric}).}
    \label{fig:double-gaussian-data-analysis-results-affine-unbend-metric-iso}
\end{figure}

In the following, we discuss how to alleviate these distortions of the manifold mappings and how to incorporate our insights into dimension reduction. The results shown in Figure~\ref{fig:double-gaussian-data-analysis-results-affine-unbend-metric-iso} indicate that the above-mentioned issues can be fully resolved by isometrizing the pullback structure.

% \todo[inline]{For all figures it is now better to have the normal and the iso versions of the figures next to each other. This will improve readability.}

% consider doing this in the appendix and just go on with the story
% [In figure] we do low-rank approximation as follows. At the base point $\mPoint\in \Real^2$ defined as the geodesic mid-point between the (Euclidean) means of the three data points inside each of the two anisotropic normal distributions, we compute the logarithmic mappings to

% \subsection{Naive manifold mappings. (1/2 page)}

% \subsection{Naive dimension reduction. (1 1/2 page)}

\subsection{Isometrized manifold mappings}
\label{sec:iso-riemannian-data-processing-mappings}

To alleviate the distortions that arise from possible discrepancies between constant speed in the $\ell^2$-sense and in the intrinsic sense, i.e., with respect to the learned pullback geometry $(\Real^\dimInd, (\cdot, \cdot)^{\diffeo})$ generated by a learned diffeomorphism $\diffeo$ on $\Real^\dimInd$, we will argue below that a simple time-change suffices. In particular, we will show below that a reparametrization of geodesics generated by a general Riemannian structure $(\Real^\dimInd, (\cdot, \cdot))$ gives rise to ``isometrized'' versions of all basic manifold mappings\footnote{We point out that the construction of the remaining manifold mappings will rely on the relationships between mappings as one would expect from Riemannian geometry, which will serve as our primary guideline for developing the theory.}. We call this framework the \emph{iso-Riemannian geometry} on $(\Real^\dimInd, (\cdot, \cdot))$ with respect to the $\ell^2$-inner product. To make the notation less dense and easier to follow for the reader, we assume in the following that the Riemannian structure $(\Real^\dimInd, (\cdot, \cdot))$ is complete. For the proofs of the results in this section we refer the reader to \Cref{app:iso-mapping-dim-1} and for the details on numerical implementations to \Cref{app:iso-mapping-dim-2}.

\paragraph{Iso-geodesics}
First, we define the \emph{iso-geodesic} mapping $\geodesic^{\iso}_{\Vector,\VectorB}:[0,1]\to \Real^\dimInd$ under $(\Real^\dimInd, (\cdot, \cdot))$ between $\Vector$ and $\VectorB\in \Real^\dimInd$ as
\begin{equation}
    \geodesic^{\iso}_{\Vector,\VectorB}(t) := \geodesic_{\Vector,\VectorB}(\timechange_{\Vector,\VectorB}(t)), \quad t\in [0,1],
    \label{eq:iso-geodesic}
\end{equation}
where $\geodesic_{\Vector,\VectorB}$ is the length-minimizing geodesic $\Vector$ and $\VectorB$ generated by $(\Real^\dimInd, (\cdot, \cdot))$, and the mapping $\timechange_{\Vector,\VectorB}:[0,1]\to[0,1]$ is a diffeomorphism, which we define by its inverse
\begin{equation}
    \timechange_{\Vector,\VectorB}^{-1}(t'):= \frac{\int_0^{t'} \|\dot{\geodesic}_{\Vector,\VectorB}(s)\|_2 ds}{\int_0^1 \|\dot{\geodesic}_{\Vector,\VectorB}(s)\|_2 ds}.
\end{equation}
The reparametrization (\ref{eq:iso-geodesic}) gives us exactly what we want, i.e., a curve with constant $\ell^2$-speed.
\begin{theorem}
\label{thm:iso-geodesic}
    For any complete Riemannian structure $(\Real^\dimInd, (\cdot, \cdot))$ on $\Real^\dimInd$, any iso-geodesic under $(\Real^\dimInd, (\cdot, \cdot))$ has constant $\ell^2$-speed.
\end{theorem}

The difference between geodesics and their isometrized versions is very noticable when comparing Figures~\ref{fig:double-gaussian-data-analysis-results-affine-unbend-metric-geo} and \ref{fig:double-gaussian-data-analysis-results-affine-unbend-metric-geo-iso}.

% \todo[inline]{[Remark that we can find tau either by (numerically) integrating  or what we found to be particularly useful if by making an approximation that the geodesic is piecewise linear between time points [t1, ...]. Then we can compute tau explicitly as. [ref to Figure]]}

% Next, we construct isometrized versions of the remainder of the manifold mappings so that relations between the iso-mappings are consistent with the original mappings.  

\paragraph{Iso-logarithms}
From the iso-geodesic mapping we can define the \emph{iso-logarithmic} mapping $\log_\Vector^{\iso}:\exp_\Vector(\mathcal{D}'_\Vector) \to \tangent_\Vector \Real^\dimInd$ under $(\Real^\dimInd, (\cdot, \cdot))$ at $\Vector \in \Real^\dimInd$ as
\begin{equation}
    \log_\Vector^{\iso}(\VectorB):= \left\{\begin{matrix}
\frac{\partial}{\partial t} \geodesic_{\Vector,\VectorB}(\timechange_{\Vector,\VectorB} (t)) \mid_{t = 0}  & \text{if } \Vector \neq \VectorB, \\
\mathbf{0} & \text{else},  \\
\end{matrix}\right.
    \label{eq:iso-log}
\end{equation}
which is consistent with the relation between the geodesic and logarithmic mapping under $(\Real^\dimInd, (\cdot, \cdot))$. 

The expression (\ref{eq:iso-log}) can be simplified further so that it becomes clear that the iso-logarithmic mapping is a rescaled logarithmic mapping whose $\ell^2$-length is equal to the arc length of the geodesic between $\Vector$ and $\VectorB$.

\begin{theorem}
\label{thm:iso-log}
    For any complete Riemannian structure $(\Real^\dimInd, (\cdot, \cdot))$ on $\Real^\dimInd$, any iso-logarithm under $(\Real^\dimInd, (\cdot, \cdot))$ satisfies
    \begin{equation}
        \log_\Vector^{\iso}(\VectorB) = \frac{\int_0^1 \|\dot{\geodesic}_{\Vector,\VectorB}(s)\|_2 ds}{\|\log_\Vector (\VectorB)\|_2 }\log_\Vector (\VectorB), \quad \VectorB \in \exp_\Vector(\mathcal{D}'_\Vector) \setminus \{\Vector\}.
        \label{eq:iso-log-simplified}
    \end{equation}
    In addition, we have that
    \begin{equation}
        \|\log_\Vector^{\iso}(\VectorB)\|_2 = \int_0^1 \|\dot{\geodesic}_{\Vector,\VectorB}(s)\|_2 ds.
        \label{eq:iso-log-length}
    \end{equation}
\end{theorem}

% \todo[inline]{Numerical approximation}

\paragraph{Iso-exponentials}
Next, we define the \emph{iso-exponential mapping} $\exp_\Vector^{\iso}:  \tangent_{\Vector}\Real^\dimInd \to \Real^\dimInd$ as
\begin{equation}
    \exp_\Vector^{\iso}(\tangentVector_\Vector) := \exp_\Vector (\vectorchange_{\tangentVector_\Vector} \tangentVector_\Vector),
    \label{eq:iso-exp}
\end{equation}
where $\vectorchange_{\tangentVector_\Vector} \geq 0$ is the minimal real number such that 
\begin{equation}
    \int_{0}^{1}\|\dot{\geodesic}_{\Vector, \exp_\Vector (\vectorchange_{\tangentVector_\Vector} \tangentVector_\Vector)}(s)\|_2 \mathrm{d}s = \|\tangentVector_\Vector\|_{2}
    \label{eq:iso-exp-criterion}
\end{equation}
holds. 

By the above construction, the iso-exponential mapping and the iso-logarithmic mapping are inverses.

\begin{theorem}
\label{thm:iso-exp}
    For any complete Riemannian structure $(\Real^\dimInd, (\cdot, \cdot))$ on $\Real^\dimInd$, we have for any $\Vector\in\Real^\dimInd$ that
    \begin{equation}
        \exp_\Vector^{\iso}(\log_\Vector^{\iso}(\VectorB)) = \VectorB, \quad \forall \VectorB \in \exp_\Vector(\mathcal{D}'_\Vector),
    \end{equation}
    and
    \begin{equation}
        \log_\Vector^{\iso}(\exp_\Vector^{\iso}(\tangentVector_\Vector)) = \tangentVector_\Vector, \quad \forall \tangentVector_\Vector \in \log_\Vector^{\iso}(\exp_\Vector(\mathcal{D}'_\Vector)).
        \label{eq:thm-iso-exp-inv}
    \end{equation}
\end{theorem}

% \todo[inline]{Numerical approximation}

\paragraph{Iso-distances}
We also define the \emph{iso-distance} mapping $\distance_{\Real^\dimInd}^{\iso}: \Real^\dimInd\times \Real^\dimInd \to \Real$ under $(\Real^\dimInd, (\cdot, \cdot))$ as
\begin{equation}
    \distance_{\Real^\dimInd}^{\iso}(\Vector, \VectorB) :=  
\int_0^1 \|\dot{\geodesic}_{\Vector,\VectorB}(s)\|_2 ds. 
\label{eq:iso-distance}
\end{equation}

However, despite the name it is important to note that we cannot expect the iso-distance mapping to be an actual metric on $\Real^\dimInd$. 

By this definition, we do recover connections with the iso-logarithmic and iso-exponential mapping that are consistent with Riemannian geometry.
\begin{theorem}
\label{thm:iso-distance}
    For any complete Riemannian structure $(\Real^\dimInd, (\cdot, \cdot))$ on $\Real^\dimInd$, we have for any $\Vector\in\Real^\dimInd$ that
    \begin{equation}
        \distance_{\Real^\dimInd}^{\iso}(\Vector, \VectorB) = \|\log_\Vector^{\iso}(\VectorB)\|_2, \quad \forall \VectorB \in \exp_\Vector(\mathcal{D}'_\Vector),
        \label{eq:thm-iso-distance-log}
    \end{equation}
    and
    \begin{equation}
        \distance_{\Real^\dimInd}^{\iso}(\Vector, \exp_\Vector^{\iso}(\tangentVector_\Vector)) = \|\tangentVector_\Vector\|_{2}, \quad \forall \tangentVector_\Vector \in \log_\Vector^{\iso}(\exp_\Vector(\mathcal{D}'_\Vector)).
        \label{eq:thm-iso-distance-exp}
    \end{equation}
\end{theorem}

\paragraph{Iso-parallel transport}
Finally, we define the \emph{iso-parallel transport} mapping $\mathcal{P}^{\iso}_{\VectorB\leftarrow \Vector} : \tangent_{\Vector} \Real^\dimInd \to \tangent_\VectorB \Real^\dimInd$ along a geodesic $\geodesic_{\Vector,\VectorB}(t)$ under $(\Real^\dimInd, (\cdot, \cdot))$ as 
\begin{equation}
    \mathcal{P}^{\iso}_{\VectorB\leftarrow \Vector}(\tangentVector_\Vector) :=  
    \Bigl(\frac{\partial \timechange_{\Vector,\VectorB}}{\partial t} (1) \frac{\partial \timechange_{\Vector,\VectorB}^{-1}}{\partial t'}(0) \Bigr)\mathcal{P}_{\VectorB\leftarrow \Vector} ( \tangentVector_\Vector).
    \label{eq:iso-parallel-transport}
\end{equation}

The expression (\ref{eq:iso-parallel-transport}) can be simplified, from which we find that it preserves $\ell^2$-length of the geodesic's velocity.

\begin{theorem}
\label{thm:iso-pt}
    For any complete Riemannian structure $(\Real^\dimInd, (\cdot, \cdot))$ on $\Real^\dimInd$, any iso-parallel transport mapping along a geodesic $\geodesic_{\Vector,\VectorB}(t)$ under $(\Real^\dimInd, (\cdot, \cdot))$ satisfies
    \begin{equation}
        \mathcal{P}^{\iso}_{\VectorB\leftarrow \Vector}(\tangentVector_\Vector) = 
    \frac{\|\dot{\geodesic}_{\Vector,\VectorB}(0)\|_2}{\|\dot{\geodesic}_{\Vector,\VectorB}(1)\|_2} \mathcal{P}_{\VectorB\leftarrow \Vector} ( \tangentVector_\Vector) = \frac{\|\log_{\Vector}(\VectorB)\|_2}{\|\log_{\VectorB}(\Vector)\|_2} \mathcal{P}_{\VectorB\leftarrow \Vector} ( \tangentVector_\Vector).
        \label{eq:iso-pt-simplified}
    \end{equation}
    In addition, we have that
    \begin{equation}
        \|\mathcal{P}^{\iso}_{\VectorB\leftarrow \Vector}(\dot{\geodesic}^{\iso}_{\Vector,\VectorB}(0))\|_2 = \|\dot{\geodesic}^{\iso}_{\Vector,\VectorB}(0)\|_2 .
        \label{eq:iso-pt-length-geo}
    \end{equation}
\end{theorem}

\begin{remark}
    Despite the above result that iso-parallel transport preserves $\ell^2$-length of geodesics along which they are transported, we do not expect this to be a general feature for any tangent vector. For this to hold, we need to assume additional properties of the Riemannian structure. For instance, assuming the structure to be locally isotropic, i.e., of the form $(\cdot, \cdot)_{\Vector} = g(\Vector)(\cdot, \cdot)_2$ for some function $g:\Real^\dimInd\to \Real_{>0}$, is sufficient, but pullback metrics cannot be expected to satisfy this constraint. 
    % For this reason we leave this for future research.
\end{remark}

% where $\isoDiffeo_\mPoint : \tangent_{\Vector} \Real^\dimInd \to \tangent_{\Vector} \Real^\dimInd$ is the diffeomorphism
% \begin{equation}
%     \isoDiffeo_\mPoint (\tangentVector_{\Vector}) := (\log_{\Vector} \circ \exp_\Vector^{\iso})(\tangentVector_\Vector) \overset{(\ref{eq:iso-exp})}{=} \vectorchange_{\tangentVector_\Vector} \tangentVector_\Vector,
%     \label{eq:l2-r-transform}
% \end{equation}
% with inverse
% \begin{equation}
%     \isoDiffeo_\mPoint^{-1} (\tangentVector_{\Vector}') := (\log_{\Vector}^{\iso} \circ \exp_\Vector)(\tangentVector_\Vector') \overset{(\ref{eq:iso-log-simplified})}{=} \frac{\int_0^1 \|\dot{\geodesic}_{\Vector,\exp_\Vector ( \tangentVector_\Vector')}(s)\|_2 ds}{\|\tangentVector_\Vector'\|_2 } \tangentVector_\Vector',
%     \label{eq:inverse-l2-r-transform}
% \end{equation}
% are (non-linear) rescalings of the input tangent vectors.

\paragraph{An alternative construction}
To see the bigger picture behind the definitions (\ref{eq:iso-geodesic}), (\ref{eq:iso-log}), (\ref{eq:iso-exp}) and (\ref{eq:iso-parallel-transport}) we define the \emph{iso-connection} or \emph{iso-covariant derivative} mapping $\nabla^{\iso}_{(\cdot)} (\cdot): \vectorfield(\Real^\dimInd) \times \vectorfield(\Real^\dimInd)\to \vectorfield(\Real^\dimInd)$ under $(\Real^\dimInd, (\cdot, \cdot))$ as 
\begin{equation}
    \nabla^{\iso}_{\tangentVectorB_\Vector} \tangentVector := \frac{1}{\|\tangentVectorB_\Vector\|_2} \nabla_{\tangentVectorB_\Vector} \|\mathcal{P}_{(\cdot) \leftarrow \Vector} \tangentVectorB_\Vector\|_2 \tangentVector ,\quad \Vector \in \Real^\dimInd, \tangentVector, \tangentVectorB \in \vectorfield(\Real^\dimInd),
    \label{eq:iso-connection}
\end{equation}
where we note that (\ref{eq:iso-connection}) indeed defines a connection, but not an affine connection as it is only 1-homogeneous in $\tangentVectorB$ rather than linear. Our claim is that this connection generates the above-mentioned manifold mappings. In other words, we claim that iso-Riemannian geometry really boils down to doing Riemannian geometry under a non-trivial connection.
% and note that this iso-connection is non-linear in both arguments -- unlike the connection.
% which can also be simplified.
% \begin{theorem}
% \label{thm:iso-cov}
%     For any complete Riemannian structure $(\Real^\dimInd, (\cdot, \cdot))$ on $\Real^\dimInd$, any iso-covariant derivative mapping along an iso-geodesic $\geodesic^{\iso}_{\Vector,\VectorB}(t) = \geodesic_{\Vector,\VectorB}(\timechange_{\Vector,\VectorB}(t))$ under $(\Real^\dimInd, (\cdot, \cdot))$ satisfies
%     \begin{equation}
%         \nabla^{\iso}_{\dot{\geodesic}^{\iso}_{\Vector,\VectorB}(t)} \tangentVector = \frac{\nabla_{\dot{\geodesic}_{\Vector,\VectorB}(\timechange_{\Vector,\VectorB}(t))} \Bigl( \|\dot{\geodesic}_{\Vector,\VectorB}\|_2 \tangentVector\Bigr)}{\|\dot{\geodesic}_{\Vector,\VectorB}(\timechange_{\Vector,\VectorB}(t))\|_2} ,\quad \tangentVector \in \vectorfield(\Real^\dimInd).
%         \label{eq:iso-cov-simplified}
%     \end{equation}
% \end{theorem}

To see this, note that the iso-covariant derivative allows us to identify vectors in different tangent spaces. In particular, we say that a vector field $\tangentVector \in \vectorfield(\geodesic)$ along a curve $\geodesic$ is \emph{iso-parallel} under $(\Real^\dimInd, (\cdot, \cdot))$ if its iso-covariant derivative vanishes, i.e.,
\begin{equation}
    \nabla^{\iso}_{\dot{\geodesic}(t)} \tangentVector = \mathbf{0}, \quad t \in [0,1].
\end{equation}
Using this definition we have that iso-geodesics are iso-parallel to themselves.
\begin{theorem}
\label{thm:iso-cov-geo}
    For any complete Riemannian structure $(\Real^\dimInd, (\cdot, \cdot))$ on $\Real^\dimInd$, any iso-geodesic under $(\Real^\dimInd, (\cdot, \cdot))$ satisfies
    \begin{equation}
        \nabla^{\iso}_{\dot{\geodesic}_{\Vector,\VectorB}^{\iso}(t)} \dot{\geodesic}_{\Vector,\VectorB}^{\iso} = \mathbf{0}.
    \end{equation}
\end{theorem}
The above statement tells us iso-geodesics are geodesics under the iso-connection, which also implies that by construction the iso-logarithmic and iso-exponential mapping are the logarithmic and exponential mappings under the iso-connection.

We also recover our definition (\ref{eq:iso-parallel-transport}) for iso-parallel transport by evaluating the following vector field at $t=1$.
\begin{theorem}
\label{thm:iso-cov-pt}
    For any complete Riemannian structure $(\Real^\dimInd, (\cdot, \cdot))$ on $\Real^\dimInd$, the vector field $\tangentVector \in \vectorfield(\geodesic_{\Vector,\VectorB})$ along the geodesic $\geodesic_{\Vector,\VectorB}$ defined as
    \begin{equation}
        \tangentVector_{\geodesic_{\Vector,\VectorB}(t)} := \frac{\|\dot{\geodesic}_{\Vector,\VectorB}(0)\|_2}{\|\dot{\geodesic}_{\Vector,\VectorB}(t)\|_2} \mathcal{P}_{\geodesic_{\Vector,\VectorB}(t)\leftarrow \Vector}(\tangentVector_\Vector), \quad \tangentVector_\Vector\in \tangent_{\Vector} \Real^\dimInd, t \in [0,1],
    \end{equation}
    is iso-parallel under $(\Real^\dimInd, (\cdot, \cdot))$. 
\end{theorem}

In other words, iso-parallel transport is parallel transport under the iso-connection.

\begin{remark}
    We finally note that there is just one notion of distance generated by a Riemannian manifold and that it is independent of the connection. So the iso-distance mapping (\ref{eq:iso-distance}) is the only mapping that is purely an analogy that follows naturally from the construction.
\end{remark}

% \todo[inline]{Note that we might not have that the length of parallel transported vectors stays the same -- unless we have a metric tensor of the form () = g(x) ()2, which is generally not the case for pullback metrics}

\subsection{Isometrized dimension reduction}
\label{sec:iso-riemannian-data-processing-low-rank}
Now that we have set up our iso-Riemannian geometry, we can use these manifold mappings to generalize algorithms for Riemannian data processing. For instance, we can formally write down an isometrized version of Algorithm~\ref{alg:l2-tangent-space-SVD} for dimension reduction, which gives Algorithm~\ref{alg:iso-l2-tangent-space-SVD}. Note that we still assume that the Riemannian structure $(\Real^\dimInd, (\cdot, \cdot))$ is complete to make the notation less dense.

\begin{algorithm}[h!]
\caption{Isometrized $\ell^2$-tangent space rank-$r$ approximation on $(\Real^\dimInd, (\cdot, \cdot))$}
\label{alg:iso-l2-tangent-space-SVD}
\begin{algorithmic}
\STATE{\textit{Initialisation}: $r\in \Natural$, $\Matrix :=[\Vector^1, \Vector^2, \ldots, \Vector^\dataPointNum]\in \Real^{\dimInd \times \dataPointNum}$, $\mPoint\in\Real^{\dimInd}$}
\STATE Compute $\log^{\iso}_\mPoint(\Matrix) := [\log^{\iso}_\mPoint(\Vector^1), \log^{\iso}_\mPoint(\Vector^2), \ldots, \log^{\iso}_\mPoint(\Vector^\dataPointNum)]\in (\tangent_\mPoint \Real^\dimInd)^\dataPointNum \cong\Real^{\dimInd \times \dataPointNum}$
\STATE Compute the SVD $\log^{\iso}_\mPoint(\Matrix) = \mathbf{U}\Sigma \mathbf{V}^\top$
\STATE Compute rank-$r$ approximation $\tangentVector_\mPoint:= [\tangentVectorComp_\mPoint^1, \tangentVectorComp_\mPoint^2, \ldots, \tangentVectorComp_\mPoint^\dataPointNum] := \mathbf{U}_r\Sigma_r \mathbf{V}_r^\top$ from the top $r$ singular values and vectors
\STATE Compute rank-$r$ approximation $\tilde{\Matrix} := [\exp^{\iso}_{\mPoint}(\tangentVectorComp_\mPoint^1), \exp^{\iso}_{\mPoint}(\tangentVectorComp_\mPoint^2), \ldots, \exp^{\iso}_{\mPoint}(\tangentVectorComp_\mPoint^\dataPointNum)]\in \Real^{\dimInd \times \dataPointNum}$
\RETURN $\tilde{\Matrix}$
\end{algorithmic}
\end{algorithm}

To argue that Algorithm~\ref{alg:iso-l2-tangent-space-SVD} solves the problem brought up at the beginning of this section, we will argue that it results in a lower value of the \emph{global approximation error}
\begin{equation}
    \|\Matrix - \exp_{\mPoint}(\isoDiffeo_\mPoint (\tangentVector_\mPoint))\|_F^2:= \sum_{\sumIndA=1}^\dataPointNum \|\Vector^\sumIndA - \exp_{\mPoint}(\isoDiffeo_\mPoint (\tangentVectorComp_\mPoint^\sumIndA))\|_2^2,
    \label{eq:global-approx-error}
\end{equation}
at some base point $\mPoint\in \Real^\dimInd$ of a data set $\Matrix :=[\Vector^1, \Vector^2, \ldots, \Vector^\dataPointNum]\in \Real^{\dimInd \times \dataPointNum}$ by a low-rank set of tangent vectors $\tangentVector_\mPoint:= [\tangentVectorComp_\mPoint^1, \tangentVectorComp_\mPoint^2, \ldots, \tangentVectorComp_\mPoint^\dataPointNum] \in (\tangent_\mPoint \Real^\dimInd)^\dataPointNum \cong\Real^{\dimInd \times \dataPointNum}$, under a certain pre-processing diffeomorphism $\isoDiffeo_{\mPoint}: \tangent_{\mPoint} \Real^\dimInd \to \tangent_{\mPoint} \Real^\dimInd$. 

\begin{remark}
\label{rem:rho-defs}
We note that the diffeomorphism $\isoDiffeo_{\Vector}$ is a convenient way to describe the errors that Algorithms~\ref{alg:l2-tangent-space-SVD}~and~\ref{alg:iso-l2-tangent-space-SVD} try to minimize, i.e., 
\begin{equation}
    \|\Matrix - \exp_{\mPoint}(\tangentVector_\mPoint)\|_F^2 \quad \text{and} \quad \|\Matrix - \exp_{\mPoint}^{\iso}(\tangentVector_\mPoint)\|_F^2,
\end{equation}
which correspond to diffeomorphisms $\isoDiffeo_{\mPoint} = \isoDiffeo_{\mPoint}^{\operatorname{id}}$ and $\isoDiffeo_{\mPoint} = \isoDiffeo_{\mPoint}^{\iso}$, where $\isoDiffeo_\mPoint^{\operatorname{id}} : \tangent_{\mPoint} \Real^\dimInd \to \tangent_{\mPoint} \Real^\dimInd$ is the identity diffeomorphism
\begin{equation}
    \isoDiffeo_\mPoint^{\operatorname{id}} (\tangentVector_{\mPoint}) := \tangentVector_\mPoint,
    \label{eq:l2-r-transform}
\end{equation}
with inverse
\begin{equation}
    (\isoDiffeo_\mPoint^{\operatorname{id}})^{-1} (\tangentVector_{\mPoint}') =  \tangentVector_\mPoint',
    \label{eq:inverse-l2-r-transform}
\end{equation}
and where $\isoDiffeo_\mPoint^{\iso} : \tangent_{\mPoint} \Real^\dimInd \to \tangent_{\mPoint} \Real^\dimInd$ is the diffeomorphism
\begin{equation}
    \isoDiffeo_\mPoint^{\iso} (\tangentVector_{\mPoint}) := (\log_{\mPoint} \circ \exp_\Vector^{\iso})(\tangentVector_\mPoint) \overset{(\ref{eq:iso-exp})}{=} \vectorchange_{\tangentVector_\mPoint} \tangentVector_\mPoint,
    \label{eq:l2-r-transform-2}
\end{equation}
with inverse
\begin{equation}
    (\isoDiffeo_\mPoint^{\iso})^{-1} (\tangentVector_{\mPoint}') = (\log_{\mPoint}^{\iso} \circ \exp_\mPoint)(\tangentVector_\mPoint') \overset{(\ref{eq:iso-log-simplified})}{=} \frac{\int_0^1 \|\dot{\geodesic}_{\mPoint,\exp_\mPoint ( \tangentVector_\mPoint')}(s)\|_2 ds}{\|\tangentVector_\mPoint'\|_2 } \tangentVector_\mPoint'.
    \label{eq:inverse-l2-r-transform-2}
\end{equation}
    
\end{remark}

To make the connection between the global approximation error (\ref{eq:global-approx-error}) to the tangent space approximation error
\begin{equation}
    \|\isoDiffeo_\mPoint^{-1} (\log_\mPoint(\Matrix)) - \tangentVector_\mPoint \|^2 := \sum_{\sumIndA=1}^\dataPointNum \|\isoDiffeo_\mPoint^{-1} (\log_\mPoint(\Vector^\sumIndA)) - \tangentVectorComp_\mPoint^\sumIndA\|_2^2,
\end{equation}
which Algorithms~\ref{alg:l2-tangent-space-SVD}~and~\ref{alg:iso-l2-tangent-space-SVD} minimize for their respective diffeomorphism $\isoDiffeo_\mPoint$, we use the following theorem, which states that the two are equivalent up to third-order if the family of matrices $\mathbf{M}_{\isoDiffeo_\mPoint}^\sumIndA \in \Real^{\dimInd\times \dimInd}$ -- each generated by data point $\Vector^\sumIndA$ and the diffeomorphism $\isoDiffeo_\mPoint$ -- as defined below is (approximately) identity for all $\sumIndA = 1, \ldots, \dataPointNum$. For the proof we refer the reader to \Cref{app:iso-mapping-dim-3}.

\begin{theorem}
\label{thm:global-approximation-error}
    Let $(\Real^\dimInd, (\cdot, \cdot))$ be any complete Riemannian structure on $\Real^\dimInd$. Furthermore, let $\mPoint\in \Real^\dimInd$ be any base point, let $\isoDiffeo_{\mPoint}: \tangent_{\mPoint} \Real^\dimInd \to \tangent_{\mPoint} \Real^\dimInd$ be any diffeomorphism, let $\Matrix:= [\Vector^1, \Vector^2, \ldots, \Vector^\dataPointNum]\in \Real^{\dimInd \times \dataPointNum}$ be any data set such that $\Vector^\sumIndA \in \exp(\mathcal{D}'_\mPoint)$ for each $\sumIndA=1,\ldots,\dataPointNum$, and let $\tangentVector_\mPoint:= [\tangentVectorComp_\mPoint^1, \tangentVectorComp_\mPoint^2, \ldots, \tangentVectorComp_\mPoint^\dataPointNum]\in (\tangent_\mPoint \Real^\dimInd)^\dataPointNum \cong \Real^{\dimInd \times \dataPointNum}$ be any approximation of $\isoDiffeo_\mPoint^{-1} (\log_\mPoint(\Matrix)):= [\isoDiffeo_\mPoint^{-1} (\log_\mPoint(\Vector^1)), \isoDiffeo_\mPoint^{-1} (\log_\mPoint(\Vector^2)), \ldots, \isoDiffeo_\mPoint^{-1} (\log_\mPoint(\Vector^\dataPointNum))]\in (\tangent_\mPoint \Real^\dimInd)^\dataPointNum \cong \Real^{\dimInd \times \dataPointNum}$.

    Then, the global approximation error (\ref{eq:global-approx-error}) reduces to
    \begin{multline}
        \|\Matrix - \exp_{\mPoint}(\isoDiffeo_\mPoint (\tangentVector_\mPoint))\|_F^2 = \sum_{\sumIndA=1}^\dataPointNum  (\isoDiffeo_\mPoint^{-1}(\log_{\mPoint}\Vector^\sumIndA) - \tangentVectorComp^\sumIndA_\mPoint)^\top\mathbf{M}_{\isoDiffeo_\mPoint}^\sumIndA(\isoDiffeo_\mPoint^{-1}(\log_{\mPoint}\Vector^\sumIndA) - \tangentVectorComp^\sumIndA_\mPoint) \\
        + \mathcal{O}(\|\isoDiffeo_\mPoint^{-1} (\log_\mPoint(\Matrix)) - \tangentVector_\mPoint \|_F^3),
        \label{eq:jacobi-bound-low-rank}
    \end{multline}
    where each $\mathbf{M}_{\isoDiffeo_\mPoint}^\sumIndA \in \Real^{\dimInd\times \dimInd}$ is given by
    \begin{equation}
       (\mathbf{M}_{\isoDiffeo_\mPoint}^\sumIndA)_{\sumIndB,\sumIndB'} := (D_{\isoDiffeo_\mPoint^{-1}(\log_{\mPoint}\Vector^\sumIndA)} \exp_{\mPoint} (\isoDiffeo_\mPoint(\cdot)) [\mathbf{e}^{\sumIndB}], D_{\isoDiffeo_\mPoint^{-1}(\log_{\mPoint}\Vector^\sumIndA)} \exp_{\mPoint} (\isoDiffeo_\mPoint(\cdot)) [\mathbf{e}^{\sumIndB'}])_2,
    \end{equation}
    where $\mathbf{e}^{\sumIndB} \in \Real^\dimInd$ is the $\sumIndB$'th standard orthonormal basis vector of $\Real^\dimInd$.
\end{theorem}

% \begin{remark}
%     The proof of Theorem~\ref{thm:global-approximation-error} follows the steps in \cite[Thm.TODO]{Diepeveen2023}, but [...]. 
% \end{remark}

Theorem~\ref{thm:global-approximation-error} tells us that to make the claim that 
\begin{equation}
    \|\Matrix - \exp_{\mPoint}(\isoDiffeo_\mPoint (\tangentVector_\mPoint))\|_F^2 \approx \|\isoDiffeo_\mPoint^{-1} (\log_\mPoint(\Matrix)) - \tangentVector_\mPoint \|_F^2,
\end{equation}
under the assumption that $\tangentVector_\mPoint \approx \isoDiffeo_\mPoint^{-1} (\log_\mPoint(\Matrix))$\footnote{which would justify deviating from the standard way of doing dimension reduction as pointed out in Remark~\ref{rem:l2-pga-instead-of-pga}} we need: (i) the linear mapping 
\begin{equation}
    \tangentVectorB \mapsto D_{\isoDiffeo_\mPoint^{-1}(\log_{\mPoint}\Vector^\sumIndA)} \exp_{\mPoint}(\isoDiffeo_\mPoint(\cdot))[\tangentVectorB]
\end{equation}
to be approximately orthogonal (so that $\mathbf{M}_{\isoDiffeo_\mPoint}^\sumIndA \approx \mathbf{I}_\dimInd$ for each $\sumIndA=1, \ldots, \dataPointNum$), and (ii) the third-order terms in (\ref{eq:jacobi-bound-low-rank}) to be small.

% For the $\mathbf{M}_{\isoDiffeo_\mPoint}^\sumIndA$ to be closer to identity when using $\isoDiffeo_{\mPoint} = \isoDiffeo_{\mPoint}^{\iso}$ rather than than $\isoDiffeo_{\mPoint} = \isoDiffeo_{\mPoint}^{\operatorname{id}}$, we need that the matrix $D_{\isoDiffeo_\mPoint^{-1}(\log_{\mPoint}\Vector^\sumIndA)} \exp_{\mPoint}$ to be closer to an orthogonal matrix for the former diffeomorphism. 

Another way of phrasing (i) is saying that $\exp_{\mPoint} (\isoDiffeo_\mPoint(\tangentVector_\mPoint))$ needs to be an approximate $\ell^2$-isometry between the tangent space and the data space for all $\tangentVector_\mPoint$ in a neighborhood of $\isoDiffeo_\mPoint^{-1}(\log_{\mPoint}\Vector^\sumIndA)$. There are two main factors that determine whether this holds: (a) geodesics having constant $\ell^2$-speed and (b) the Riemannian structure $(\Real^\dimInd, (\cdot, \cdot))$ having zero curvature -- as geodesics diverge for negative curvature and vice versa for positive curvature. The former is true for $\isoDiffeo_{\mPoint} = \isoDiffeo_{\mPoint}^{\iso}$, but not necessarly for $\isoDiffeo_{\mPoint} = \isoDiffeo_{\mPoint}^{\operatorname{id}}$, whereas the latter is true if the underlying Riemannian structure is a Euclidean pullback metric, i.e., $ (\cdot, \cdot) =  (\cdot, \cdot)^\diffeo$ for a diffeomorphism $\diffeo:\Real^\dimInd\to \Real^\dimInd$.

% []
% which is the case by construction as iso-Riemannian geometry is designed to enhance correspondence of distances in the tangent space to distances in data space. 

In addition, when (i) holds for all points in our data set, we also expect that (ii) holds as higher order derivatives of $\tangentVector_\mPoint \mapsto \exp_{\mPoint} (\isoDiffeo_\mPoint(\tangentVector_\mPoint))$ will be small, which results in very small higher order terms in (\ref{eq:jacobi-bound-low-rank}) since we also have that $\tangentVector_\mPoint \approx \isoDiffeo_\mPoint^{-1} (\log_\mPoint(\Matrix))$. 

In other words, we expect that for a low-rank approximation through Algorithm~\ref{alg:iso-l2-tangent-space-SVD} (with high enough rank) under an Euclidean pullback structure on $\Real^\dimInd$, we have
\begin{equation}
    \tangentVector_\mPoint \approx \isoDiffeo_\mPoint^{-1} (\log_\mPoint(\Matrix)) \quad \text{and} \quad \|\Matrix - \exp_{\mPoint}^{\iso}(\tangentVector_\mPoint)\|_F^2 \approx \|\log_\mPoint^{\iso}(\Matrix) - \tangentVector_\mPoint \|^2,
\end{equation}
from which follows that $\|\Matrix - \exp_{\mPoint}^{\iso}(\tangentVector_\mPoint)\|_F^2$ is also small, which is in line with observations (see Figure~\ref{fig:double-gaussian-data-analysis-results-affine-unbend-metric-pga-iso}). In contrast, there might still be a (possibly large) discrepancy between the global approximation and tangent space error in the case of Algorithm~\ref{alg:l2-tangent-space-SVD}, which is also in line with observations (see Figure~\ref{fig:double-gaussian-data-analysis-results-affine-unbend-metric-pga}).

\begin{remark}
    We note that a curvature and isometry correction scheme to dimension reduction -- analogous to the scheme proposed in \cite{Diepeveen2023} -- can be constructed based on the rewritten global approximation error in Theorem~\ref{thm:global-approximation-error}. However, for the Euclidean pullback setting we are mainly interested in in this work, we noticed hardly any difference after already using iso-manifold mappings\footnote{We occasionally even observed that performance deteriorated as numerical inaccuracies in computing $\mathbf{M}_{\isoDiffeo_\mPoint}^\sumIndA$ arose from compounded approximation errors -- those introduced by isometrized mappings as well as from the finite difference approximation of the differential.}. Having that said, from \cite{Diepeveen2023} and subsequent work \cite{Chew2025} we do expect that in the case of pulling back a Riemannian structure with curvature, we can no longer ignore the $\mathbf{M}_{\isoDiffeo_\mPoint}^\sumIndA$. However, a proper study into how to handle both isometry and curvature at the same time is beyond the scope of this work.
\end{remark}

% [Remark other methods
% \cite{Chew2025}
% \cite{ho2013nonlinear}
% ]
\section{Regular normalizing flow-based Riemannian data processing of real-valued data}
\label{sec:regular-nfs}

Next, we continue with addressing the second question at the end of Section~\ref{sec:intro}: 
\begin{center}
    \emph{Can recent developments -- including overshadowed regular linear architectures and expressive non-volume-preserving innovations -- enable diffeomorphisms to model complex manifolds without sacrificing regularity?}
\end{center}
However, before going into detail on how to parametrize and train diffeomorphisms without sacrificing regularity, we again aim to give the reader some intuition as to why lacking regularity can potentially be an issue in the setting of multimodal data distributions. In particular, 
% to see why even isometrizing cannot save every learned Riemannian structure, 
we will consider learning pullback geometry with the (overly expressive) diffeomorphisms and the training scheme proposed in \cite{Diepeveen2024a} on multimodal data. While the methodology proposed in \cite{Diepeveen2024a} showed empirical effectiveness on the multimodal MNIST data set\footnote{despite being primarily designed for unimodal distributions}, we argue that architectures used -- affine coupling flows \cite{Dinh2017} and neural spline flows \cite{Durkan2019} -- may fail to preserve necessary geometric properties in multimodal settings. To illustrate this, we will again consider problems arising in learned geodesics and learned non-linear low rank approximation.

% \subsection{Irregular normalizing flows.}

\begin{figure}[h!]
    \centering
    \begin{subfigure}[b]{0.49\linewidth}
        \centering
        \includegraphics[width=\linewidth]{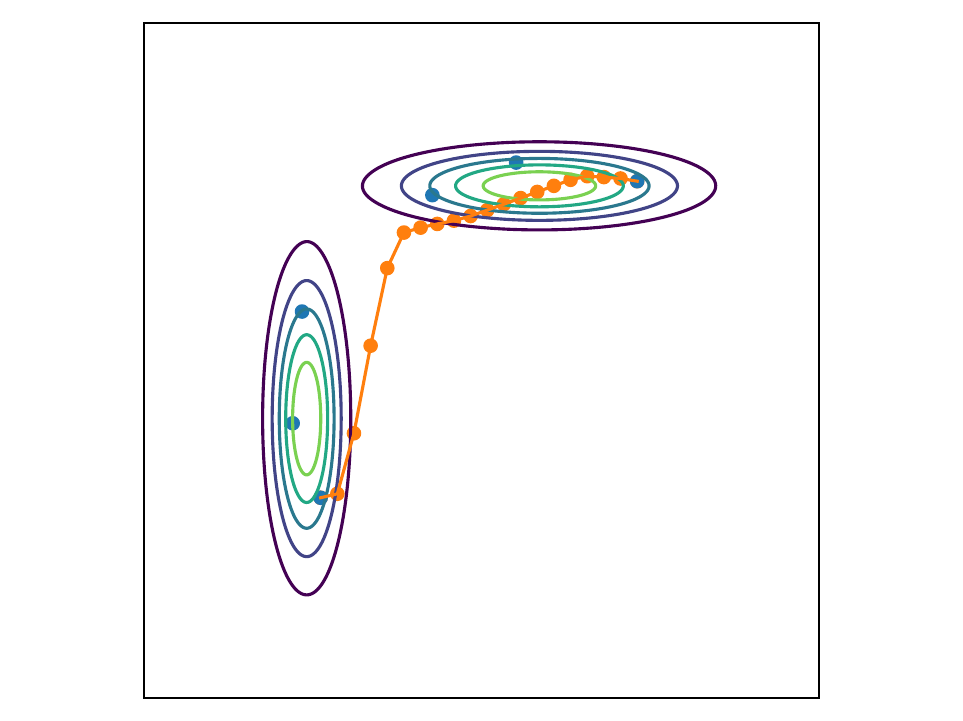}
        \caption{Expressive pullback geodesic}
        \label{fig:double-gaussian-data-analysis-results-affine-metric-geo}
    \end{subfigure}
    \hfill
    \begin{subfigure}[b]{0.49\linewidth}
        \centering
        \includegraphics[width=\linewidth]{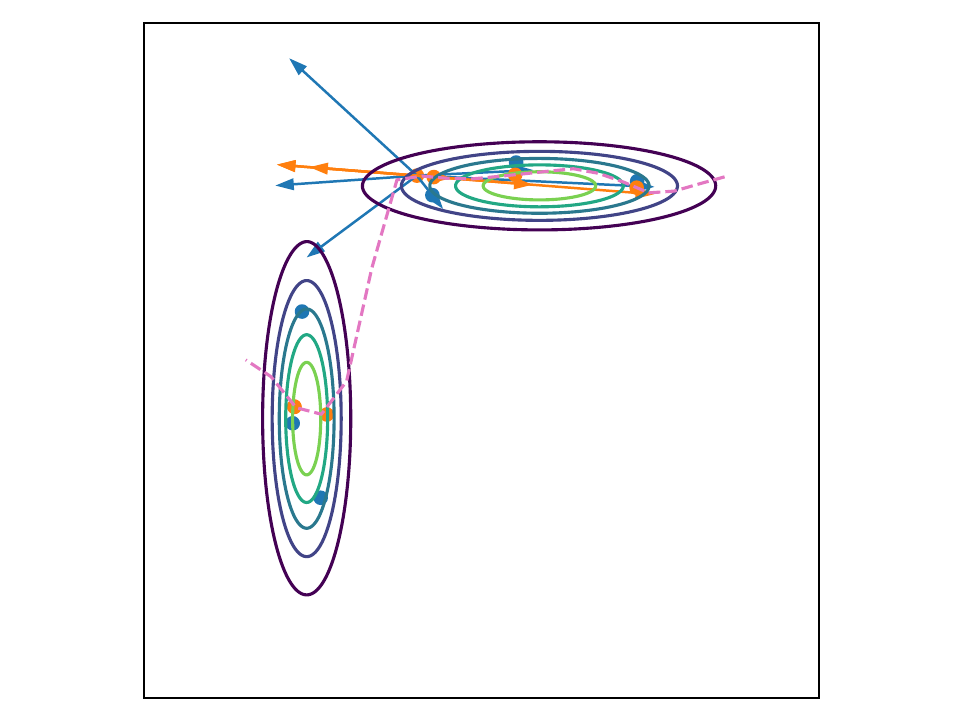}
        \caption{Expressive pullback rank-1 approximation}
        \label{fig:double-gaussian-data-analysis-results-affine-metric-pga}
    \end{subfigure}
    \caption{Both geodesic interpolation and non-linear rank-1 approximation under a learned pullback structure \cite{Diepeveen2024a} on $\Real^2$ do not behave as expected, which is detrimental for interpolate interpretability and rank-1 reconstruction fairness.}
    \label{fig:double-gaussian-data-analysis-results-affine-metric}
\end{figure}

We will train a pullback structure as proposed in \cite{Diepeveen2024a} on samples from the bimodal normal distribution from Figure~\ref{fig:double-gaussian-data-analysis-results-toy-metric-geo} and validate the learned pullback geometry on the same six validation data points in blue as before. The results shown in Figure~\ref{fig:double-gaussian-data-analysis-results-affine-metric} indicate that several things are going wrong:

\paragraph{Incorrect geodesics} 
Because there is little to no data in between the two modes, there will always be an ambiguity how the geodesic travels from one mode to the other. In Figure~\ref{fig:double-gaussian-data-analysis-results-affine-metric-geo} we see that in the learned metric geodesics starting from the top right mode leave this mode in a natural way, but enter the lower left mode from the side rather than from the top (which would be the more natural way as shown in Figure~\ref{fig:double-gaussian-data-analysis-results-toy-metric-geo}). In other words, we found a pullback metric, but not the one we wanted and one can imagine as the data distribitions become higher dimensional, there are more ways to transition from one mode to another. So this problem is expected to be worse as dimension increases.

\paragraph{Incorrect dimension reduction} 
Having incorrect geodesics directly breaks down downstream tasks such as dimension reduction as this relies on the assumption that the data manifold coincides with a geodesic subspace. We see in Figure~\ref{fig:double-gaussian-data-analysis-results-affine-metric-pga} that this geodesic subspace assumption definitely does not hold as the learned manifold in pink is a particularly poor approximation of the data manifold we expect to find, i.e., one that looks more like the geodesic from Figure~\ref{fig:double-gaussian-data-analysis-results-toy-metric-geo}. In addition, approximations of the data points on the data manifold naturally vary wildly in quality: whereas the projections on the manifold for the top right mode are reasonable, the converse holds for the approximations in the lower left mode. This will be detrimental for the interpretability and when considering applications pertaining fairness as the error for one part of the data set is (unnecessarily) much larger than for the another part.

\begin{remark}
    Note that isometrizing the manifold mappings as we have done in the previous section will not resolve anything when the underlying geometry generates inherently incorrect manifold mappings.
\end{remark}

\begin{figure}[h!]
    \centering
    \begin{subfigure}[b]{0.49\linewidth}
        \centering
        \includegraphics[width=\linewidth]{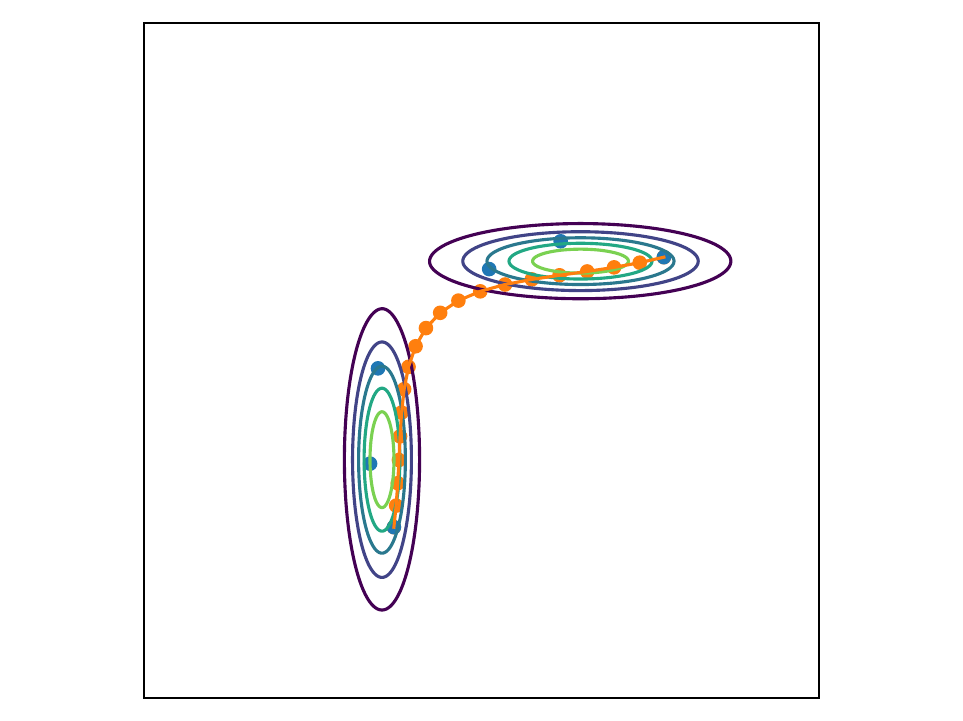}
        \caption{Regular pullback geodesic}
        \label{fig:double-gaussian-data-analysis-results-additive-tanh-metric-geo}
    \end{subfigure}
    \hfill
    \begin{subfigure}[b]{0.49\linewidth}
        \centering
        \includegraphics[width=\linewidth]{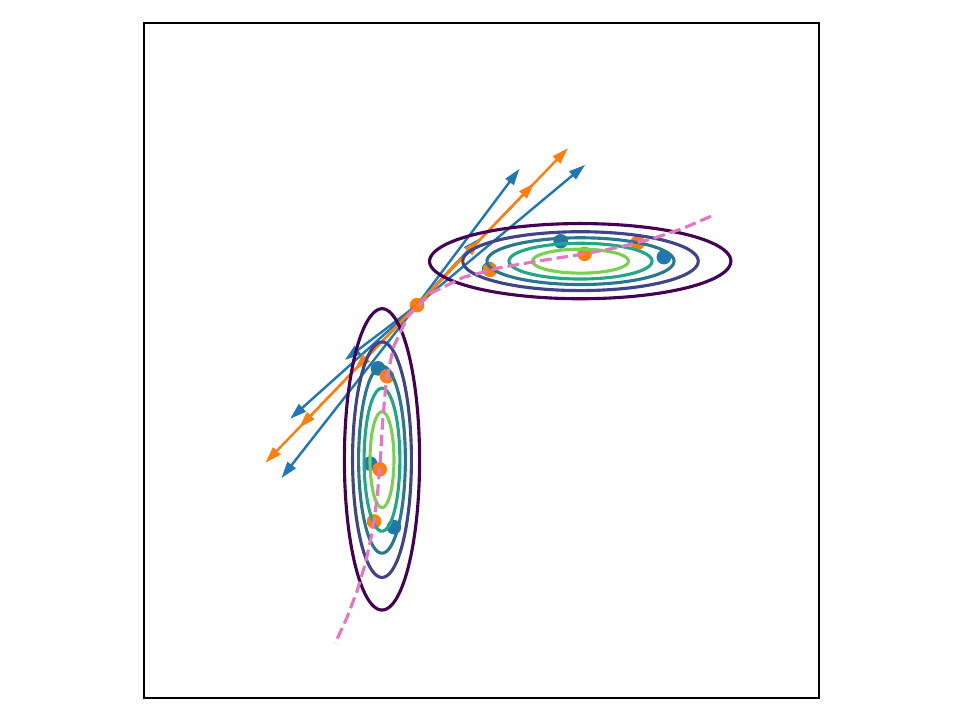}
        \caption{Regular pullback rank-1 approximation}
        \label{fig:double-gaussian-data-analysis-results-additive-tanh-metric-pga}
    \end{subfigure}
    \caption{Neither geodesic interpolation nor non-linear rank-1 approximation under a more regular learned pullback structure on $\Real^2$ have learned an incorrect geometry, in contrast to the initial pullback structure (see Figure~\ref{fig:double-gaussian-data-analysis-results-affine-metric}).}
    \label{fig:double-gaussian-data-analysis-results-additive-tanh-metric}
\end{figure}

In the following, we will discuss how to get in a better position for constructing data-driven pullback geometry through using more regular diffeomorphisms and standard normalizing flow training, while incorporating recent advances. The results in Figure~\ref{fig:double-gaussian-data-analysis-results-additive-tanh-metric} (which we consistently achieve) indicate that our odds of learning the right way of transitioning between modes increase through parametrizing our diffeomorphisms with more regular layers under this new constraint.

\subsection{Parametrization}
The reason why more regular diffeomorphisms could alleviate incorrect geodesics, is because by limiting expressivity we force the network to learn the simplest way to transition from one mode to the next, i.e., the one with the least twists and turns. However, we do want to have enough expressivity to learn suitable pullback geometry that captures the complex data manifold. We will argue below that we can do this by adding more invertible linear layers, while making the non-linearities more regular.

% \paragraph{Euclidean data.}
First, we note that the claims in \cite{Diepeveen2024,Diepeveen2024a} -- that the diffeomorphism generating the pullback geometry has to be locally $\ell^2$-isometric on the data support -- might be too pessimistic given the results in Sections~\ref{sec:iso-riemannian-data-processing-mappings} and \ref{sec:iso-riemannian-data-processing-low-rank}. In particular, we have seen that we do not expect any issues with the downstream data processing as long as geodesics have constant $\ell^2$-speed and the exponential mapping is a local $\ell^2$-isometry between parts of data space where the data resides and the corresponding parts of the tangent space. This is still the case if we deviate from the additive coupling-based framework proposed in \cite{Dinh2014} by giving up on the diffeomorphism to be volume-preserving -- while maintaining that the mapping $\Vector\mapsto \log(|\det(D_{\Vector} \diffeo)|)$ is constant --  and using non-linearities with bounded derivatives. 

By this line of reasoning we propose to consider diffeomorphisms $\diffeo_\networkParams:\mathbb{E}\to\mathbb{E}$ between Euclidean spaces $\mathbb{E}$, e.g., $\mathbb{E}=\Real^\dimInd$ for vector data and $\mathbb{E}=\Real^{c \times h\times w}$ for images, that are compositions of $L\in \Natural$ diffeomorphic blocks
\begin{equation}
    \diffeo_\networkParams = \diffeoB_{\networkParams^L} \circ \ldots \diffeoB_{\networkParams^1},
\end{equation}
where each diffeomorphism $\diffeoB_{\networkParams^\sumIndC}:\mathbb{E}\to\mathbb{E}$ boils down to a mapping of the form
\begin{equation}
    \diffeoB_{\networkParams^\sumIndC}(\Vector):= (\diffeoC_{\networkParams_1^\sumIndC} \circ \diffeoD_{\networkParams_2^\sumIndC})(\Vector), \quad \networkParams^\sumIndC:= (\networkParams_1^\sumIndC, \networkParams_2^\sumIndC),
\end{equation}
where the mapping $\diffeoC_{\networkParams_1^\sumIndC}:\mathbb{E}\to\mathbb{E}$ is an additive coupling mapping \cite{Dinh2014} and $\diffeoD_{\networkParams_2^\sumIndC}:\mathbb{E}\to\mathbb{E}$ is an invertible linear mapping. For the following it is convenient to assume that the diffeomorphisms $\diffeoC_{\networkParams^\sumIndC}$ are of the form
\begin{equation}
    \diffeoC_{\networkParams_1^\sumIndC}(\Vector):=  \left\{\begin{matrix}
\Vector_\sumIndB & \text{if } \sumIndB \in J, \\
\Vector_\sumIndB + f_{\networkParams_1^\sumIndC}(\mathbf{m}_J \odot \Vector)_{\sumIndB} & \text{if } \sumIndB \notin J, \\
\end{matrix}\right.
\label{eq:additive-coupling}
\end{equation}
where $f_{\networkParams_1^\sumIndC}:\mathbb{E}\to\mathbb{E}$ is a neural network with learnable activation functions\footnote{That is, for every node in every hidden layer we assume an activation function with its own weights.} 
\begin{equation}
    \sigma(x):= \sum_{\sumIndD=1}^N a_\sumIndD \tanh(x)^\sumIndD,
\end{equation}
and $\mathbf{m}_J \in \mathbb{E}$ is a binary mask
\begin{equation}
    \mathbf{m}_J:= \left\{\begin{matrix}
1 & \text{if } \sumIndB \in J, \\
0 & \text{if } \sumIndB \notin J, \\
\end{matrix}\right. 
\end{equation}
so that $f_{\networkParams_1^\sumIndC}(\mathbf{m}_J \odot \Vector)$ is effectively a function of the entries $\Vector_\sumIndB$ with $\sumIndB \in J$.

\begin{remark}
    Note that this choice of parametrization does what we set out to do. That is, the invertible linear mappings are not necessarily volume-preserving but will always have a constant determinant, while the (volume-preserving) additive coupling layers have bounded derivatives -- whose sizes we can keep small by promoting small coefficients $a_\sumIndD$ -- due to the choice of activation.
\end{remark}

% \begin{remark}
%     We note that [our choice for parametrization is somewhat different than the architecture used in \cite{Dinh2014}, in which the authors only used the additive layers with fixed activation. The parametrization above gives more flexibility -- expressivity due to the learnable activation functions and non-volume preserving diffeomorphisms with constant log determinant due to the linear layers -- while maintaining regularity -- differentiability and bounded derivatives due to the learnable activation functions.] 
% \end{remark}

For specific data types, there are particular choices that we observed to work well. These correspond or are inspired by both overshadowed and popular architectures from the normalizing flow literature.

\paragraph{Vector data}
The remaining details when considering vector data are the coupling mapping $f_{\networkParams_1^\sumIndC}:\Real^\dimInd\to \Real^\dimInd$, the index set $J$, and the invertible linear layer $\diffeoD_{\networkParams_2^\sumIndC}:\Real^\dimInd\to \Real^\dimInd$.
For the neural network $f_{\networkParams_1^\sumIndC}$ we observed that feed forward networks typically suffice. For the index set $J$, any way to split the $\dimInd$ entries into two groups should work, although we use 
$$
J = \{\sumIndB \in [\dimInd] \mid \sumIndB \operatorname{mod} 2 = 0\},
$$
to stay in line with the image setting (discussed below). Finally, for the invertible linear mapping $\diffeoD_{\networkParams_2^\sumIndC}$ we observed that learnable normalization followed by an orthogonal matrix parametrized through the Householder decomposition \cite{Tomczak2016} tends to perform well, where latter is a decomposition as the a product of reflection matrices 
$$
    \mathbf{I} - \frac{1}{\|\mathbf{v}\|_2^2} \mathbf{v} \mathbf{v}^\top,\quad \text{with learnable } \mathbf{v} \in \Real^\dimInd.
$$
\begin{remark}
    It is worth highlighting that this combination for the linear mapping also underlies the additional layers in the Glow architecture \cite{Kingma2018}. In particular, the proposed linear layer can be seen as an activation normalization followed by a $1\times 1$ convolution with an orthogonal matrix.
\end{remark}

\paragraph{Image data}
The remaining details when considering image data are also the coupling mapping $f_{\networkParams_1^\sumIndC}:\Real^{c \times h\times w} \to \Real^{c \times h\times w}$, the index set $J$, and the invertible linear layer $\diffeoD_{\networkParams_2^\sumIndC}:\Real^{c \times h\times w} \to \Real^{c \times h\times w}$. 

For the neural network $f_{\networkParams_1^\sumIndC}$ we observed that convolutional neural networks typically suffice. For the index set $J$ we want to encode that only local information in the image should be used to update entries. To do this we use a checkerboard indexing
$$
J = \{(\sumIndB_c, \sumIndB_h, \sumIndB_w) \in [c] \times [h] \times [w] \mid \sumIndB_h + \sumIndB_w \operatorname{mod} 2 = 0\}.
$$
Additionally, we share activation functions in the image height and width dimensions, i.e., only for different channels we learn different weights for the $\tanh$ summands.

Finally, for the invertible linear mapping $\diffeoD_{\networkParams_2^\sumIndC}$ we observed that activation normalization followed by two invertible convolutions, i.e., mappings of the form (\ref{eq:additive-coupling}) where the coupling is a 2D convolution with some kernel $\mathbf{K}\in \Real^{c \times c \times \kappa \times \kappa}$ with kernel size $\kappa\in \Natural$ and where the first convolution updates all $\sumIndB\in J$ and the second convolution updates the all $\sumIndB\notin J$.

\begin{remark}
    We note that we deviate from the $1\times 1$ convolution from the Glow architecture \cite{Kingma2018} to enhance expressivity.
\end{remark}

\subsection{Training}
Now that we have proposed to use a more regular parametrization, our next claim is that we can greatly simplify the loss used in \cite{Diepeveen2024a}. The original loss is composed of a negative log-likelihood and two regularization terms that promote the diffeomorphism to be volume-preserving and $\ell^2$-isometric on the data support, and is used to train an anisotropic normalizing flow. The loss we propose is the standard normalizing flow loss with weight decay:
\begin{equation}
    \mathcal{L}(\networkParams) := \mathbb{E}_{\stoVector \sim \density_{\text{data}}}\left[-\log \density_{\networkParams}(\stoVector)\right] + \frac{\lambda}{2}\|\networkParams\|_F^2,
    \label{eq:loss}
\end{equation}
where $\density_{\networkParams} :\mathbb{E}\to \Real$ given by
\begin{equation}
    \density_{\networkParams}(\Vector) := \frac{1}{\sqrt{(2\pi)^\dimInd}}e^{-\frac{1}{2} \|\diffeo_{\networkParams}(\Vector)\|^2_F } |\det(D_{\Vector} \diffeo_{\networkParams})|.
\end{equation}

To motivate this loss, we need to argue why we can switch back to a loss that is more like the standard normalizing flow loss despite the empirical evidence from \cite{Diepeveen2024a} that the standard normalizing flow loss has its problems. First, we note that if one insists on having a locally $\ell^2$-isometric diffeomorphism, it is necessary to use an anisotropic base distribution. However, as we have argued above, we can give up on being volume-preserving as long as we have a constant determinant and bounded derivatives. We note that the constant determinant is guaranteed by the parametrization and that the latter is promoted by the parametrization and can be enforced by the weight decay term\footnote{unlike the standard normalizing flows compared to in \cite{Diepeveen2024a}, which were parametrized by affine coupling flows}. So there is potentially no need to stick to the anisotropic normalizing flow training scheme from \cite{Diepeveen2024a}. Instead, it should suffice to learn pullback geometry through minimizing (\ref{eq:loss}), which is supported by the results in Figure~\ref{fig:double-gaussian-data-analysis-results-additive-tanh-metric}.

\begin{remark}
    We also note that the proposed loss (\ref{eq:loss}) does no longer have regularization terms that depend on the support of the data, which we have seen to  be possibly problematic in the case of multimodal data (see Figure~\ref{fig:double-gaussian-data-analysis-results-affine-metric}).
\end{remark}

\section{Numerical Experiments}
\label{sec:numerics}

\begin{figure}[h!]
    \centering
    \begin{subfigure}[b]{0.49\linewidth}
        \centering
        \includegraphics[width=\linewidth]{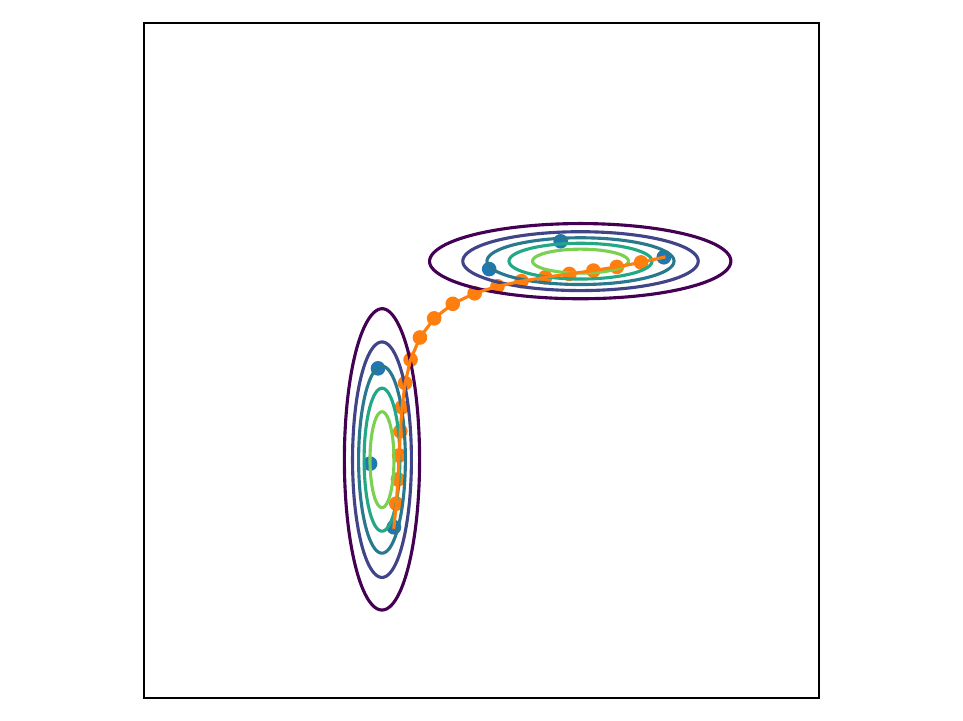}
        \caption{Isometrized pullback geodesic}
        \label{fig:double-gaussian-data-analysis-results-iso-additive-tanh-metric-geo}
    \end{subfigure}
    \hfill
    \begin{subfigure}[b]{0.49\linewidth}
        \centering
        \includegraphics[width=\linewidth]{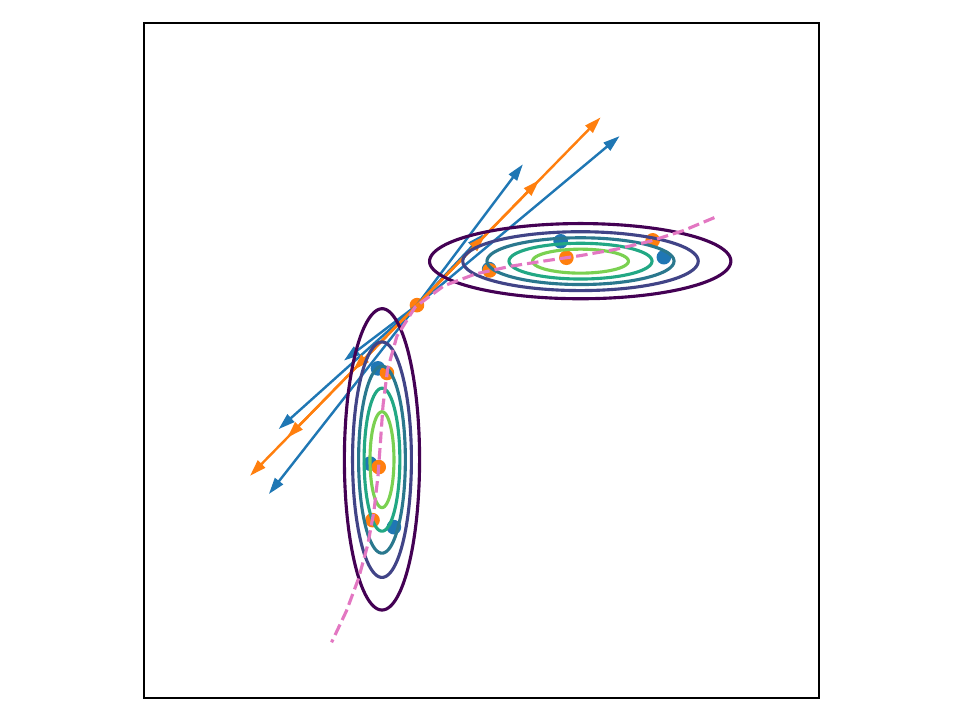}
        \caption{Isometrized pullback rank-1 approximation}
        \label{fig:double-gaussian-data-analysis-results-iso-additive-tanh-metric-pga}
    \end{subfigure}
    \caption{Isometrizing geodesic interpolation and non-linear rank-1 approximation under a more regular learned pullback structure on $\Real^2$ yields additional improvement over the non-isometrized versions (see Figure~\ref{fig:double-gaussian-data-analysis-results-additive-tanh-metric}).}
    \label{fig:double-gaussian-data-analysis-results-iso-additive-tanh-metric}
\end{figure}

In the previous sections we have seen how isometrizing or using more regular parametrizations can help resolve challenges that both general and learned pullback metrics can suffer from. Unsurprisingly, when combining best practices, i.e., using isometrized regular learned pullback geometry, the results for regular learned pullback geometry (see Figure~\ref{fig:double-gaussian-data-analysis-results-additive-tanh-metric}) can be improved further, as shown in Figure~\ref{fig:double-gaussian-data-analysis-results-iso-additive-tanh-metric}. 
% However, we note that the difference -- between using a pullback structure and using it isometrized -- is more pronounced in the case of the modeled pullback metric compared to the regular learned pullback metric. This is visually clear by comparing the improvement from Figure~\ref{fig:double-gaussian-data-analysis-results-affine-unbend-metric} to Figure~\ref{fig:double-gaussian-data-analysis-results-affine-unbend-metric-iso} and from Figure~\ref{fig:double-gaussian-data-analysis-results-additive-tanh-metric} to Figure~\ref{fig:double-gaussian-data-analysis-results-iso-additive-tanh-metric}, but 
This can be made more concrete when considering the relative RMSEs of the rank-1 approximation $\tangentVector_\mPoint:= [\tangentVectorComp_\mPoint^1, \tangentVectorComp_\mPoint^2, \ldots, \tangentVectorComp_\mPoint^\dataPointNum] \in (\tangent_\mPoint \Real^\dimInd)^\dataPointNum \cong\Real^{\dimInd \times \dataPointNum}$ at the Riemannian barycentre $\mPoint\in \Real^\dimInd$
\begin{equation}
    \operatorname{Low-rank \; rel-RMSE} := 
    % \frac{\|\Matrix - \exp^\diffeo_{\mPoint}(\isoDiffeo_\mPoint (\tangentVector_\mPoint^\sumIndA))\|_F}{\|\Matrix - \mPoint\|_F} := 
    \sqrt{\frac{\frac{1}{\dataPointNum}\sum_{\sumIndA=1}^\dataPointNum\|\Vector^\sumIndA - \exp^\diffeo_{\mPoint}(\isoDiffeo_\mPoint (\tangentVectorComp_\mPoint^\sumIndA))\|_2^2}{\frac{1}{\dataPointNum}\sum_{\sumIndA=1}^\dataPointNum \|\Vector^\sumIndA - \mPoint\|_2^2}},
\end{equation}
under Algorithms~\ref{alg:l2-tangent-space-SVD} and \ref{alg:iso-l2-tangent-space-SVD} -- $\isoDiffeo_\mPoint = \isoDiffeo_\mPoint^{\operatorname{id}}$ and $\isoDiffeo_\mPoint = \isoDiffeo_\mPoint^{\iso}$ (see Remark~\ref{rem:rho-defs}) -- in Table~\ref{tab:low-rank-errors-double-gaussians}. 

\begin{table}[h!]
    \centering
    \caption{Caption}
    \label{tab:low-rank-errors-double-gaussians}
    \begin{tabular}{lccc}
\toprule
Metric & Figure & Iso & Rank-1 rel-RMSE \\
\midrule
Modeled pullback & \ref{fig:double-gaussian-data-analysis-results-affine-unbend-metric} & \xmark & 0.1741 \\
Modeled pullback & \ref{fig:double-gaussian-data-analysis-results-affine-unbend-metric-iso} & \cmark & 0.0606 \\
Regular learned pullback & \ref{fig:double-gaussian-data-analysis-results-additive-tanh-metric} & \xmark & 0.1146 \\
Regular learned pullback & \ref{fig:double-gaussian-data-analysis-results-iso-additive-tanh-metric} & \cmark & 0.0868 \\
\bottomrule
\end{tabular}
\end{table}

Having said that in the Table we see that the error drop is much larger for the modeled case compared to the learned case, one could still say that isometrizing gave noticeable performance boost. 

Motivated by the above discussion, we finish this work by addressing the question:
\begin{center}
    \emph{When learning pullback geometry through the proposed regular normalizing flow parametrization and training scheme, does isometrizing the learned geometry systematically give an improvement in performance?}
\end{center}
We aim to get insight into this question through looking at both synthetic data (hemisphere) and real data (MNIST). 

Besides the low-rank rel-RMSE, it will be useful to quantify the extent to which non-isometrized geodesics have approximately constant $\ell^2$-speed. We propose to consider the relative RMSE between the geodesics from the Riemannian barycentre $\mPoint\in \Real^\dimInd$ to all data points
\begin{equation}
    \operatorname{Geodesic \; rel-RMSE} :=  \sqrt{\frac{\frac{1}{\dataPointNum}\sum_{\sumIndA=1}^\dataPointNum \frac{1}{m+1}\sum_{\sumIndB=0}^{m} \| \geodesic^{\diffeo}_{\mPoint, \Vector^\sumIndA}(\sumIndB/m) - \geodesic^{\diffeo, \iso}_{\mPoint, \Vector^\sumIndA}(\sumIndB/m)\|_2^2}{\frac{1}{\dataPointNum}\sum_{\sumIndA=1}^\dataPointNum \|\Vector^\sumIndA - \mPoint\|_2^2}},\quad m\in \Natural.
\end{equation}
For the modeled pullback structure we have that $\operatorname{geodesic \; rel-RMSE} = 0.2428$, whereas for the regular learned pullback structure we have that $\operatorname{geodesic \; rel-RMSE} = 0.0156$.

Finally, it will also be insightful to see which data points tend to contribute most to the above-mentioned relative errors. We expect that these are the points that are furthest away from the barycentre $\mPoint$. We will test our hypothesis by considering the point clouds in $\Real^2$ given by 
\begin{equation}
    (\|\Vector^\sumIndA - \mPoint\|_2, \sqrt{\frac{1}{m+1}\sum_{\sumIndB=0}^{m} \| \geodesic^{\diffeo}_{\mPoint, \Vector^\sumIndA}(\sumIndB/m) - \geodesic^{\diffeo, \iso}_{\mPoint, \Vector^\sumIndA}(\sumIndB/m)\|_2^2}), \quad \sumIndA=1, \ldots, \dataPointNum,
\end{equation}
and
\begin{equation}
    (\|\Vector^\sumIndA - \mPoint\|_2, \|\Vector^\sumIndA - \exp^\diffeo_{\mPoint} (\tangentVectorComp_\mPoint^\sumIndA)\|_2 -  \|\Vector^\sumIndA - \exp^{\diffeo, \iso}_{\mPoint}(\tangentVectorComp_\mPoint^\sumIndA)\|_2), \quad \sumIndA=1, \ldots, \dataPointNum.
\end{equation}

% [in the previous sections we have already done several numerical experiments on the double gaussian data to argue and visualize how the proposed measures improve upon naive ways of using and constructing Riemannian geometry]. In the following we will shift our focus towards how isometrizeed Riemannian geometry and regular parametrizations combine. Once again, we will consider the synthetic data we have used before, but will also use a more realistic image data set.

% [do already a plot here that we get better performance for the base data set when we use both isometrizing and regular diffeos + say that we will consider this in more detail in experiments with higher dimensional data manifolds -- one synthetic (hemisphere) and one real (mnist)]

% \todo[inline]{We need an appendix where we refer to all the training details for all numerics (so also the toy ones we have been using throughout the paper). We could also do a table [table for the errors from the low-rank approximation]}

For all experiments in this section, 
% details on the data sets are provided in [appendix], 
detailed training configurations are provided in \Cref{app:training-details}.

% [also mention that we evaluate the geometry on a validation set rather than on the training data.]

\subsection{Hemisphere}

\begin{figure}[h!]
    \centering
    \begin{subfigure}[b]{0.49\linewidth}
        \centering
        \includegraphics[width=\linewidth]{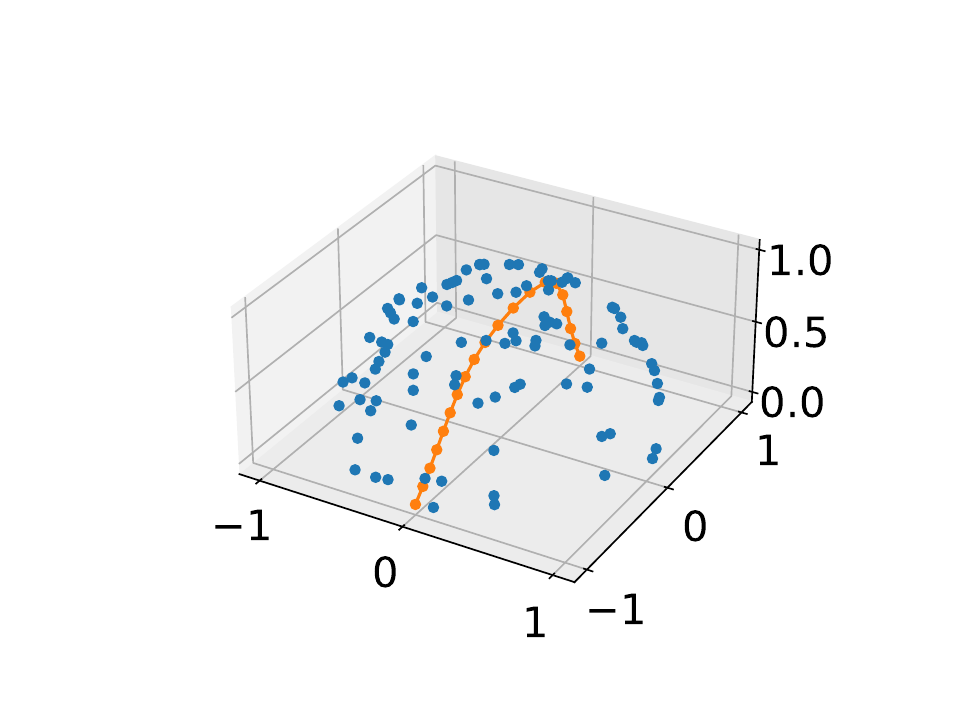}
        \caption{Pullback geodesic}
    \end{subfigure}
    \hfill
    \begin{subfigure}[b]{0.49\linewidth}
        \centering
        \includegraphics[width=\linewidth]{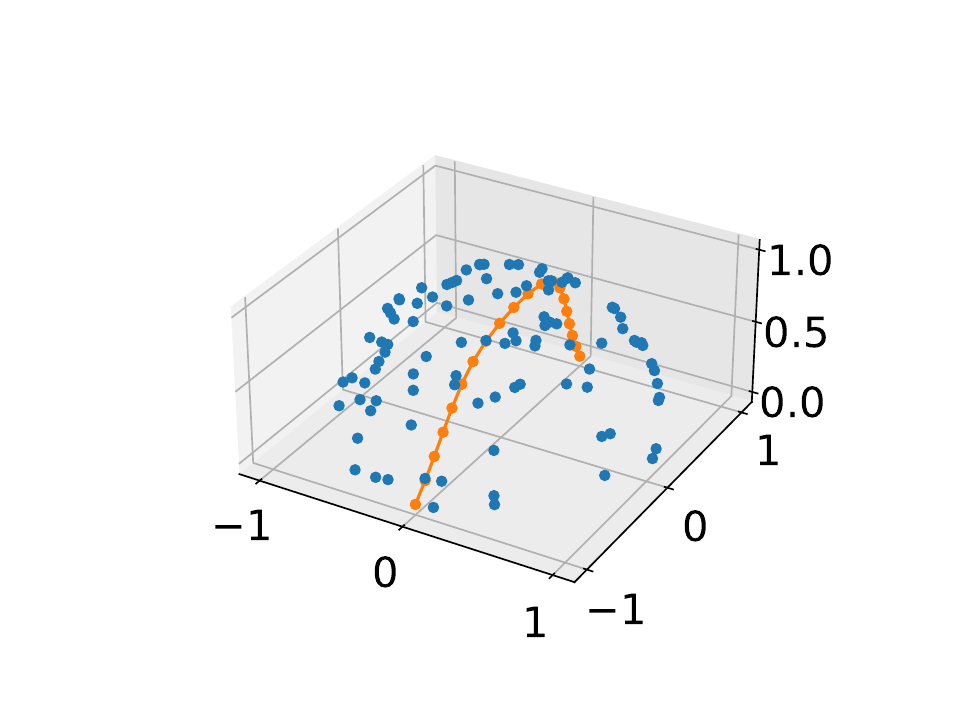}
        \caption{Isometrized pullback geodesic}
    \end{subfigure}
    \caption{The learned geodesic between antipodal points of the hemisphere (orange) has a slightly lower $\ell^2$-speed on the far side compared to the near side, which is why the spacing between the iso-geodesic interpolates on the near side is slightly larger.}
    \label{fig:hemi-sphere-geodesics}
\end{figure}

\begin{figure}[h!]
    \centering
    \begin{subfigure}[b]{0.49\linewidth}
        \centering
        \includegraphics[width=\linewidth]{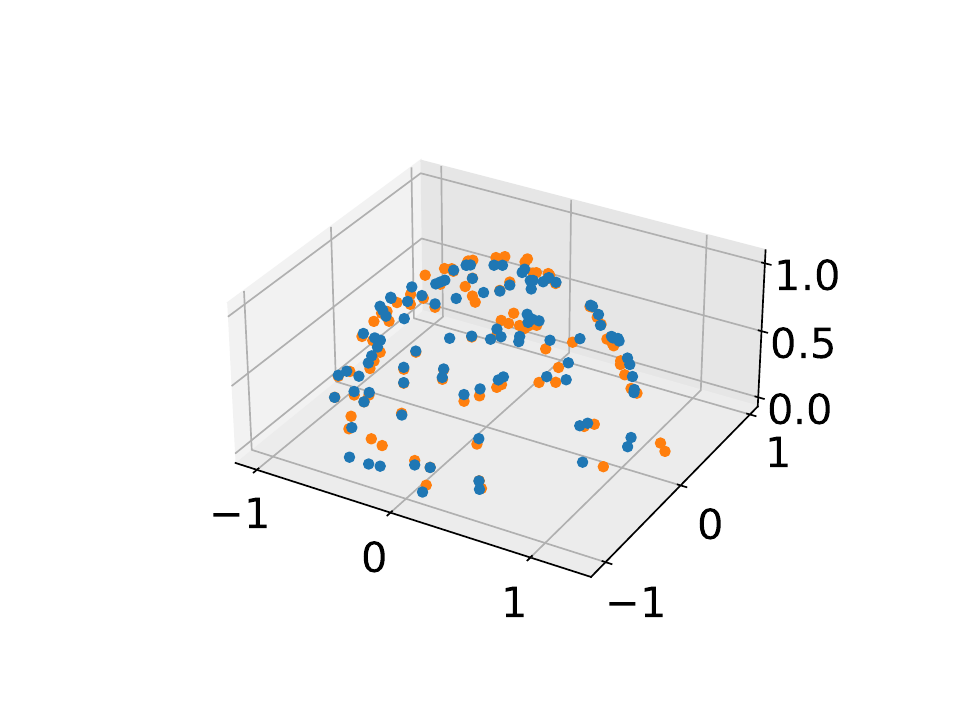}
        \caption{Pullback rank-2 approximation}
    \end{subfigure}
    \hfill
    \begin{subfigure}[b]{0.49\linewidth}
        \centering
        \includegraphics[width=\linewidth]{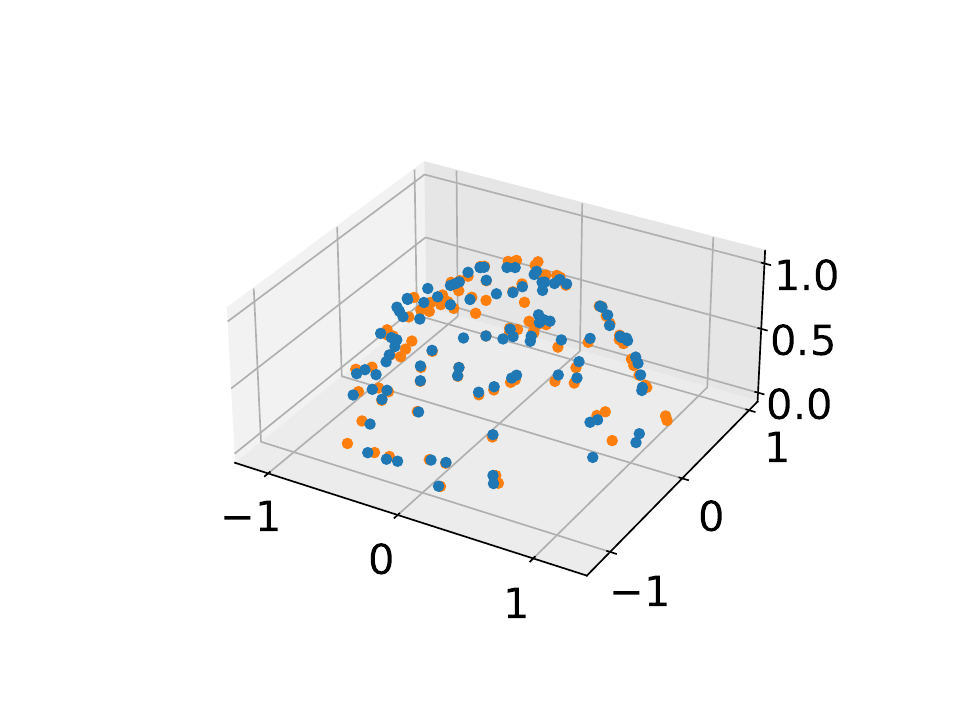}
        \caption{Isometrized pullback rank-2 approximation}
    \end{subfigure}
    \caption{Low-rank approximations based on both the learned pullback geometry and its iso-Riemannian counterpart effectively capture the data manifold (orange), though isometrization yields slightly more accurate representations.}
    \label{fig:hemi-sphere-low-rank}
\end{figure}

\begin{figure}[h!]
    \centering
    \begin{subfigure}[b]{0.49\linewidth}
        \centering
        \includegraphics[width=\linewidth]{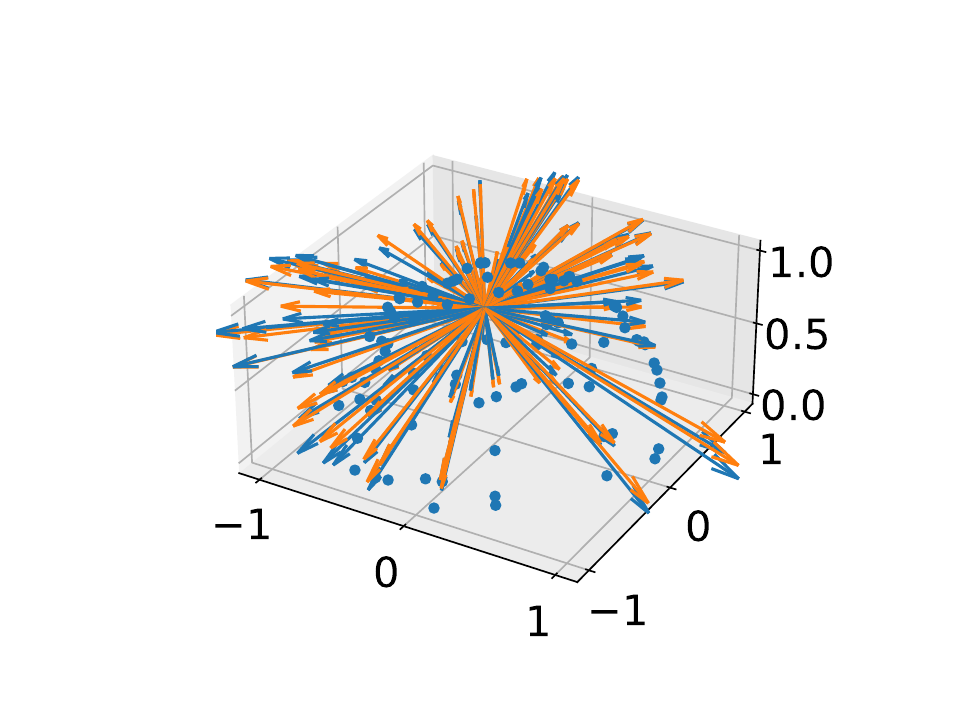}
        \caption{Pullback logarithmic mappings}
    \end{subfigure}
    \hfill
    \begin{subfigure}[b]{0.49\linewidth}
        \centering
        \includegraphics[width=\linewidth]{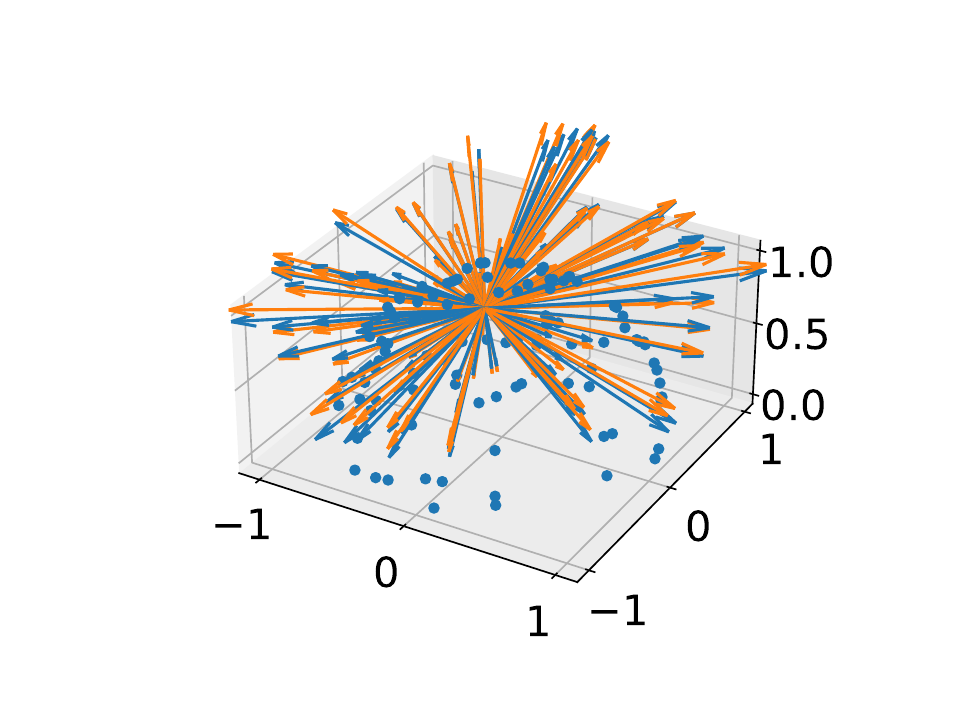}
        \caption{Isometrized pullback logarithmic mappings}
    \end{subfigure}
    \caption{Without isometrization, the learned logarithmic mappings (blue) and the rank-2 approximation (orange) introduce distortions that make data on one side of the tangent space appear farther away than data on the opposite side.}
    \label{fig:hemi-sphere-tangent-vectors}
\end{figure}

\begin{figure}[h!]
    \centering
    \begin{subfigure}[b]{0.49\linewidth}
        \centering
        \includegraphics[width=\linewidth]{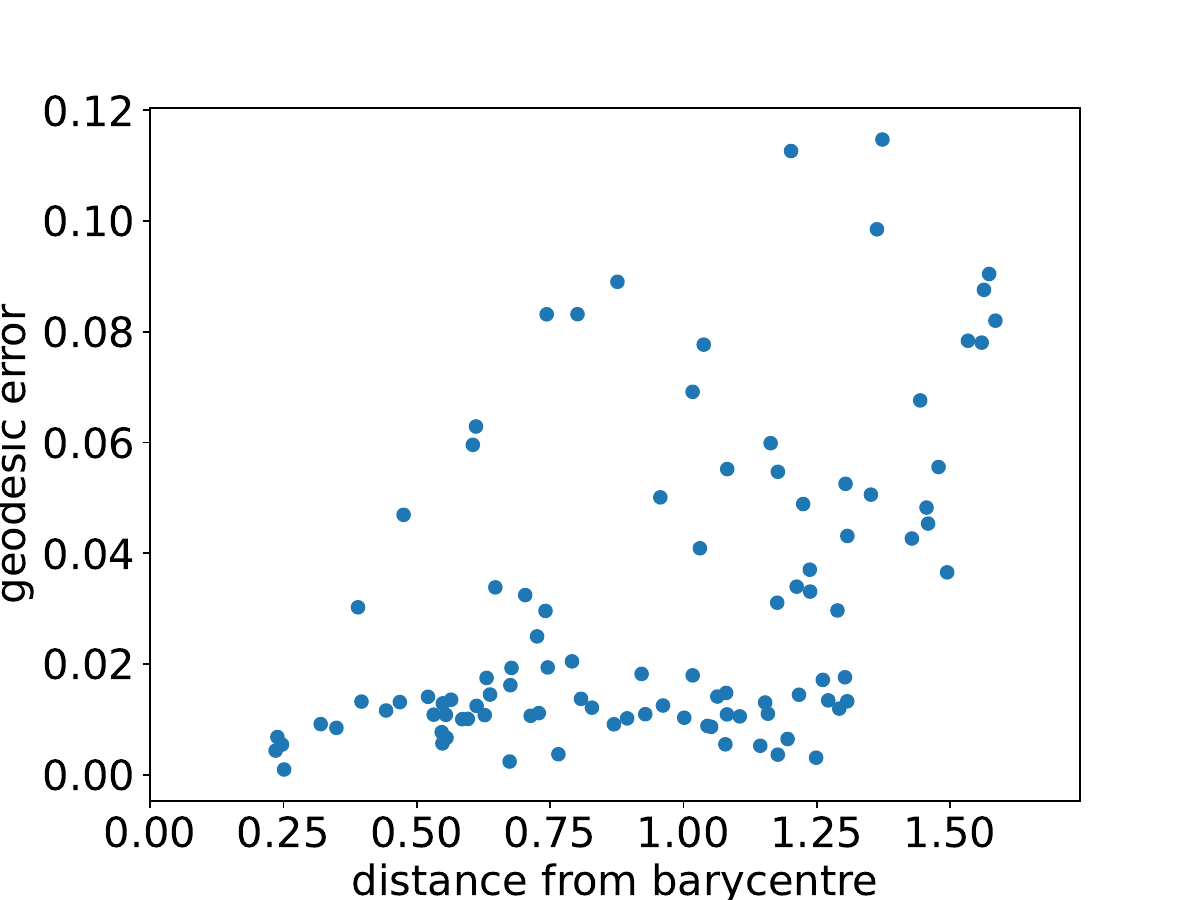}
        \caption{}
        \label{fig:hemisphere-data-analysis-results-additive-tanh-metric-geo-errors}
    \end{subfigure}
    \hfill
    \begin{subfigure}[b]{0.49\linewidth}
        \centering
        \includegraphics[width=\linewidth]{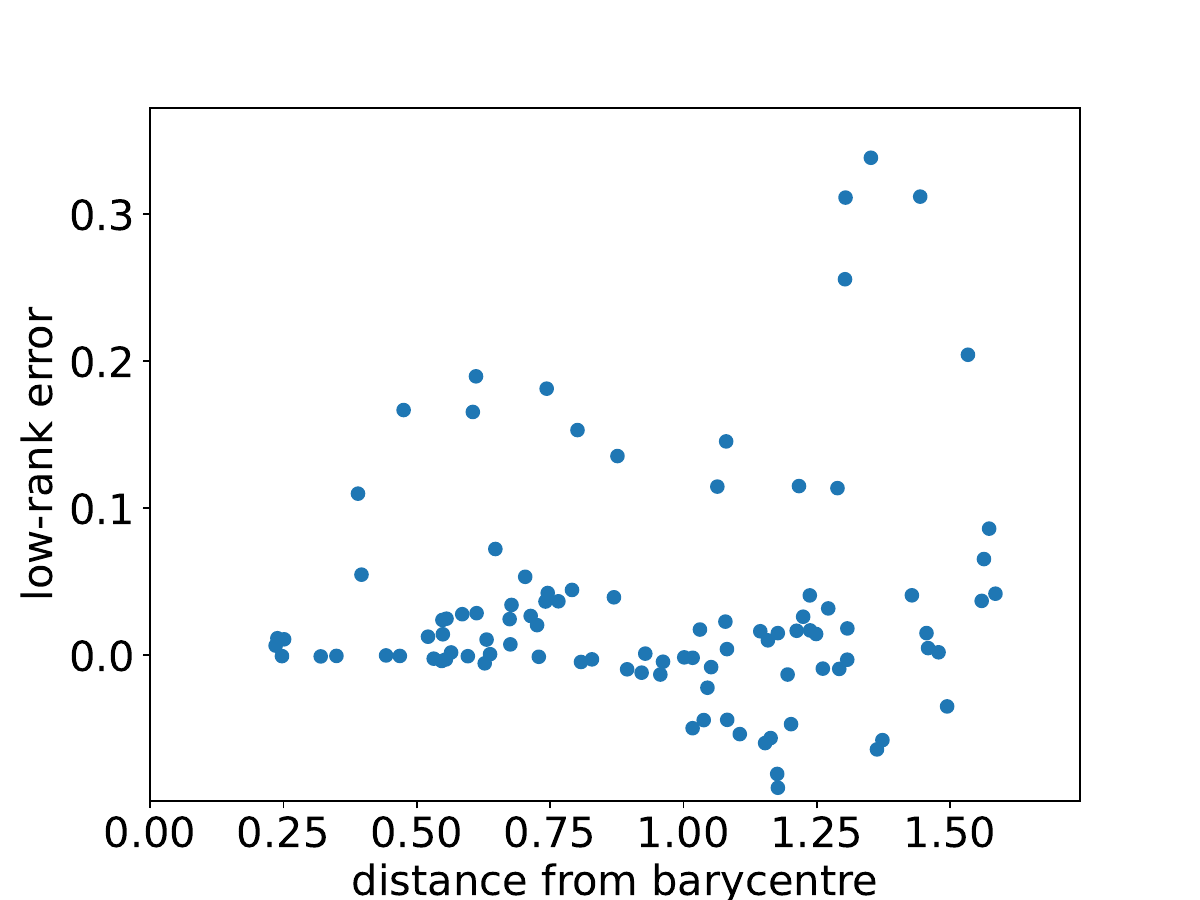}
        \caption{}
        \label{fig:hemisphere-data-analysis-results-iso-additive-tanh-metric-low-rank-errors}
    \end{subfigure}
    \caption{For the majority of data points, the choice between learned pullback geometry and isometrized geometry makes little difference; however, as points move farther from the data barycentre, isometrization becomes increasingly significant.}
    \label{fig:hemisphere-data-analysis-results-additive-tanh-metric-errors}
\end{figure}

When evaluating the learned pullback geometry on $\dataPointNum=100$ hemisphere validation data points, we find $\operatorname{geodesic \; rel-RMSE} = 0.0409$, and for rank-2 approximation we find $\operatorname{low-rank \; rel-RMSE} = 0.1682$ under Algorithm~\ref{alg:l2-tangent-space-SVD} and $\operatorname{low-rank \; rel-RMSE} = 0.1153$ under Algorithm~\ref{alg:iso-l2-tangent-space-SVD}. These results are also visually in line with what we see. That is, the geodesics in Figure~\ref{fig:hemi-sphere-geodesics} are almost identical and the low-rank approximations in Figure~\ref{fig:hemi-sphere-low-rank} have learned the same subspace -- although the approximation of the data is visibly better for the isometrized approximation. Having that said, we would like to point out that when considering the (iso-)logarithmic mappings (and tangent space approximations thereof) in Figure~\ref{fig:hemi-sphere-tangent-vectors}, we see a very distinct difference, which tells us that the improvement in low-rank approximation comes from a lack of isometry of the exponential mapping. This effect also becomes more prevalent as the data points are further from the barycentre, which can also be observed from the worst-case discrepancies in Figure~\ref{fig:hemisphere-data-analysis-results-additive-tanh-metric-errors}. In other words, isometrizing was essential in faithful data processing for this data set.

\clearpage

\subsection{MNIST}
\begin{figure}[h!]
    \centering
    \begin{subfigure}{\linewidth}
        \centering
        \includegraphics[width=0.8\linewidth]{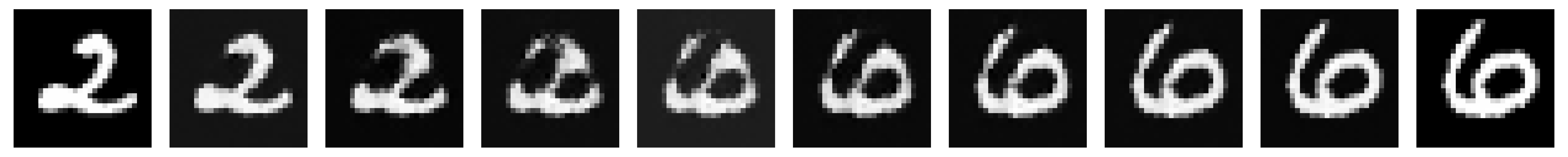}
        \caption{Pullback geodesic}
    \end{subfigure}
    \begin{subfigure}{\linewidth}
        \centering
        \includegraphics[width=0.8\linewidth]{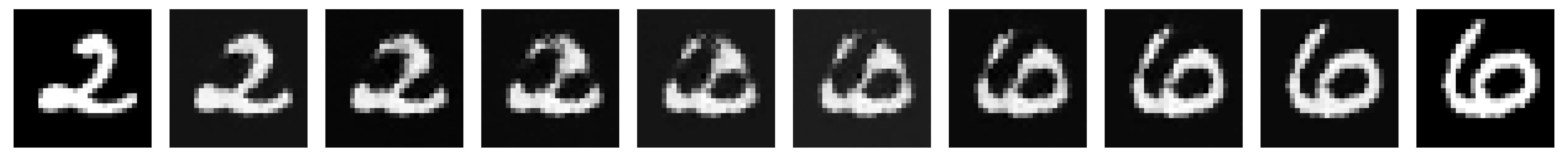}
        \caption{Isometrized pullback geodesic}
    \end{subfigure}
    \begin{subfigure}{\linewidth}
        \centering
        \includegraphics[width=0.8\linewidth]{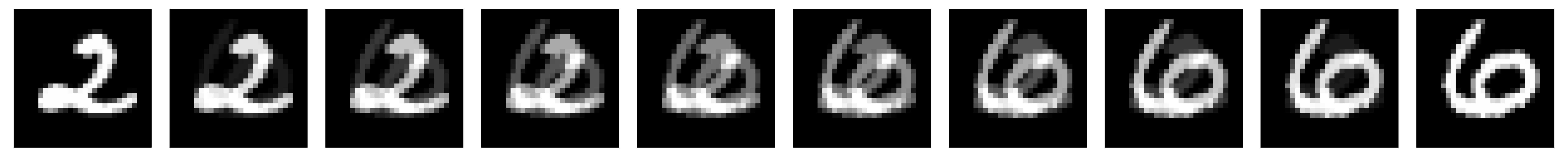}
        \caption{Linear interpolation}
    \end{subfigure}
    \caption{The learned geodesic and iso-geodesic between `2' and `6' are nearly indistinguishable, yet both produce a notably more natural interpolation path compared to linear interpolation.}
    \label{fig:mnist-geodesics}
\end{figure}

\begin{figure}[h!]
    \centering
    \begin{subfigure}{\linewidth}
        \centering
        \includegraphics[width=0.8\linewidth]{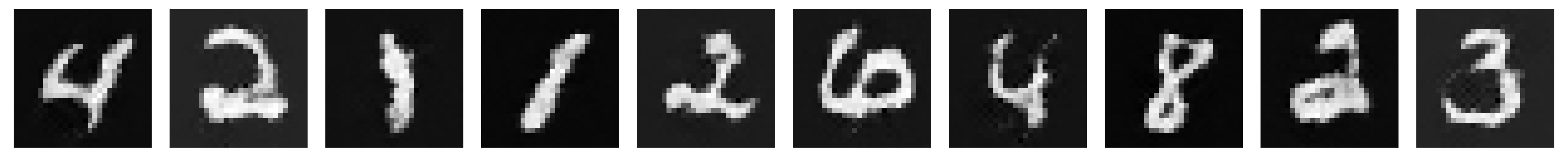}
        \caption{Pullback rank-20 approximation}
    \end{subfigure}
    \begin{subfigure}{\linewidth}
        \centering
        \includegraphics[width=0.8\linewidth]{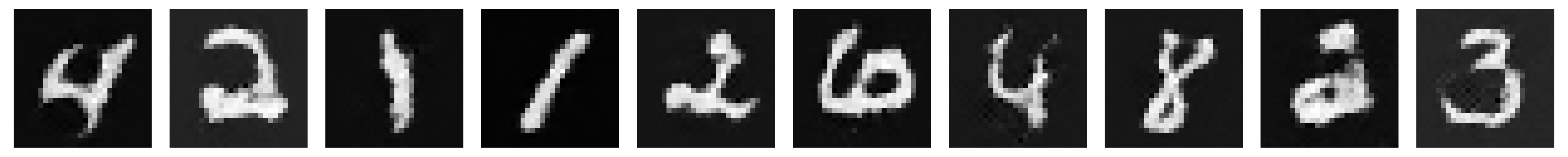}
        \caption{Isometrized pullback rank-20 approximation}
    \end{subfigure}
    \begin{subfigure}{\linewidth}
        \centering
        \includegraphics[width=0.8\linewidth]{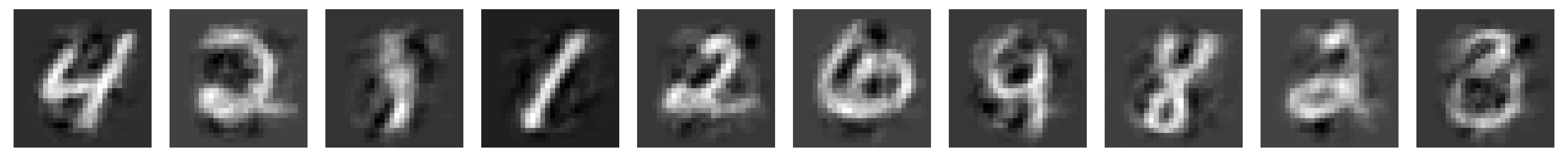}
        \caption{Linear rank-20 approximation (PCA)}
    \end{subfigure}
    \begin{subfigure}{\linewidth}
        \centering
        \includegraphics[width=0.8\linewidth]{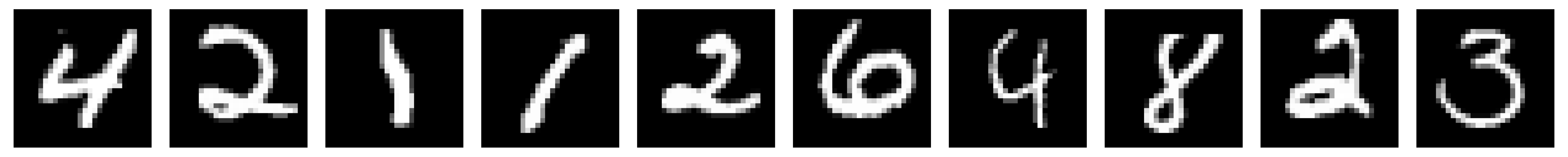}
        \caption{Ground truth}
    \end{subfigure}
    \caption{Low-rank approximations based on both the learned pullback geometry and its iso-Riemannian counterpart effectively capture the data manifold and yield similarly accurate representations -- especially compared to linear low-rank approximation.}
    \label{fig:mnist-low-rank}
\end{figure}

\begin{figure}[h!]
    \centering
    \begin{subfigure}[b]{0.49\linewidth}
        \centering
        \includegraphics[width=\linewidth]{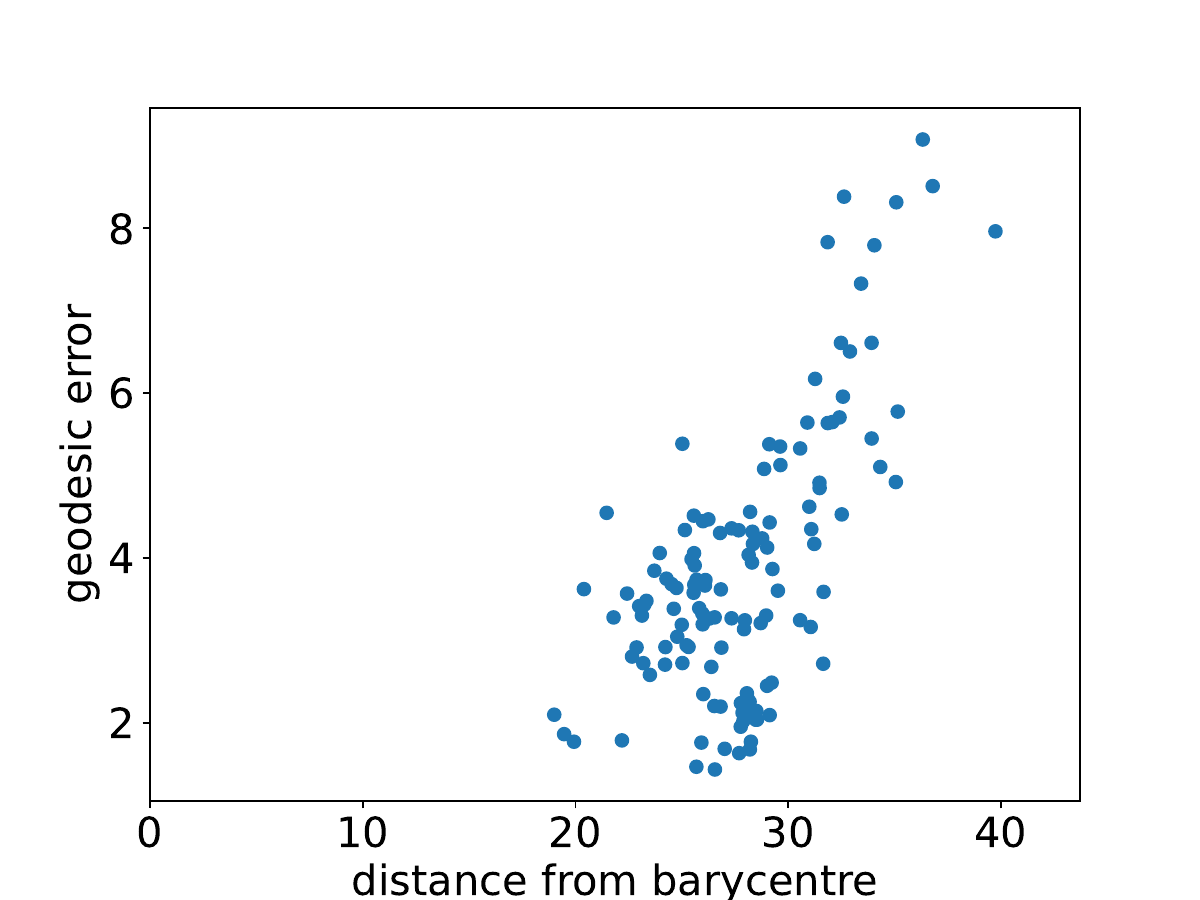}
        \caption{}
        \label{fig:mnist-data-analysis-results-additive-tanh-metric-geo-errors}
    \end{subfigure}
    \hfill
    \begin{subfigure}[b]{0.49\linewidth}
        \centering
        \includegraphics[width=\linewidth]{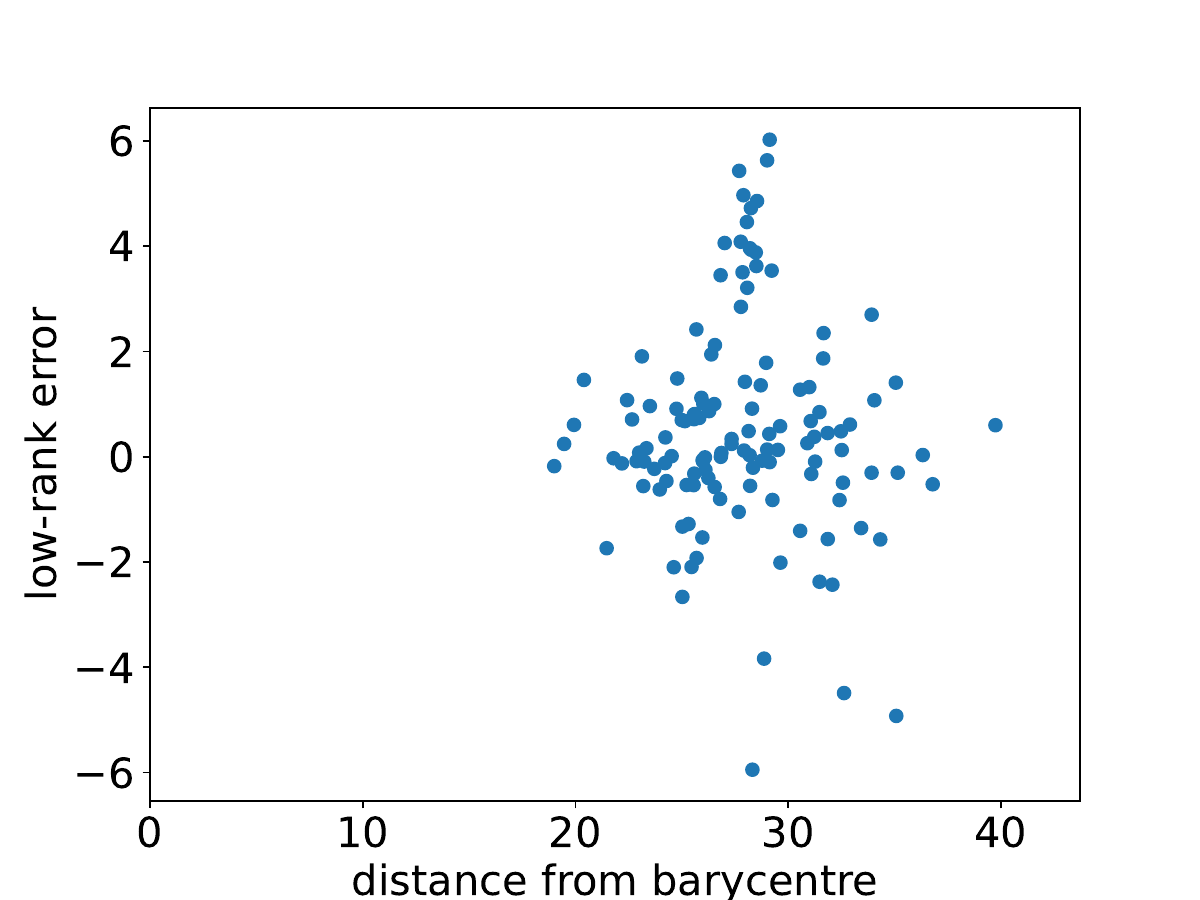}
        \caption{}
        \label{fig:mnist-data-analysis-results-iso-additive-tanh-metric-low-rank-errors}
    \end{subfigure}
    \caption{As data points lie farther from the learned barycentre, the disparity between the learned pullback geometry and isometrized geometry intensifies for geodesic interpolation -- a stark contrast to low-rank approximation, where both geometries exhibit nearly equivalent performance.}
    \label{fig:mnist-data-analysis-results-additive-tanh-metric-errors}
\end{figure}

When evaluating the learned pullback geometry on $\dataPointNum=128$ MNIST validation data images, we find $\operatorname{geodesic \; rel-RMSE} = 0.1497$, and for rank-20 approximation we find $\operatorname{low-rank \; rel-RMSE} = 0.5147$ under Algorithm~\ref{alg:l2-tangent-space-SVD} and $\operatorname{low-rank \; rel-RMSE} = 0.5042$ under Algorithm~\ref{alg:iso-l2-tangent-space-SVD}. These results are also visually in line with what we see. That is, the geodesics in Figure~\ref{fig:mnist-geodesics} differ slightly in the middle frames, whereas the low-rank approximations in Figure~\ref{fig:mnist-low-rank} are almost identical\footnote{Although it is clear that we have learned a non-linear metric when comparing to what we would get in the standard $\ell^2$-inner product, i.e., linear interpolation and linear rank-20 approximation (PCA).}. Upon closer inspection in Figure~\ref{fig:mnist-data-analysis-results-additive-tanh-metric-errors}, we see that the geodesic error seems to increase systematically as the distance from the barycentre to the data end point increases -- more so compared to the hemisphere data. However, this does not seem to have an effect on the dimension reduction. So overall, isometrizing still has a noticeable effect, but is less essential in faithful data processing for this data set.

\section{Conclusions}
\label{sec:conclusions}

This work addresses key challenges in data-driven pullback geometry -- specifically distortions from non-isometric manifold mappings and modeling inaccuracies stemming from diffeomorphism irregularity. To resolve these, we introduce (i) isometrization of the Riemannian structure, formalized as iso-Riemannian geometry, and argue for (ii) regular normalizing flows that maintain expressivity while simplifying training. Empirical results demonstrate that these approaches synergistically mitigate geometric distortions and modeling errors, with combined application yielding superior performance compared to isolated implementations.

Within the broader scope of data processing, this work can be seen as an attempt to refocus efforts towards pullback geometry-based data-driven Riemannian geometry, emphasizing regularity, expressivity, and iso-Riemannian geometry.

% With this investigation, we hope to further push the focus of data-driven Riemannian geometry towards pullback geometry and its development towards regularity, expressivity and iso-Riemannian geometry. In particular, within the broader scope of data processing, we believe that the proposed mathematical framework has important implications on how to construct and post-process data-driven Riemannian geometry and has important implications for handling data in general.

% In this work we aimed to address problems in data-driven pullback geometry pertaining distortions due to a lack of isometry and modeling errors due to a lack of regularity. We proposed to address these challenges through isometrizing the Riemannian structure, which gave rise to iso-Riemannian geometry, and through using regular yet expressive normalizing flows, which also greatly simplified the training scheme. We showcased that these two approaches effectively alleviate distortions and reduce modeling errors, and additionally work in a complementary fashion, i.e., when combined give superior performance.


\subsubsection*{Acknowledgments}
%\todo[inline]{@Deanna: please write a blurb on how this was funded}

Both authors were partially supported by NSF DMS 2408912. The authors would also like to thank Joyce Chew for fruitful discussions.

% \clearpage
\bibliographystyle{plain}
\bibliography{Bibliography}

\appendix
\section{An illustrative Riemannian structure for a bimodal Gaussian distribution}
% \label{app:illustrative-geometries}

% \subsection{Data-driven pullback geometry for a bimodal Gaussian distribution.}
\label{app:illustrative-pullback}
% \todo[inline]{Write out}

The diffeomorphism $\diffeo:\Real^2\to \Real^2$ used for the pullback metric in Figures~\ref{fig:double-gaussian-data-analysis-results-toy-metric-geo} to \ref{fig:double-gaussian-data-analysis-results-affine-unbend-metric-iso} is defined as
\begin{equation}
    \diffeo(\Vector):= \diffeoB ( \mathbf{R} \Vector), 
\end{equation}
where the matrix $\mathbf{R}\in \Real^{2\times 2}$ is the rotation matrix
\begin{equation}
    \mathbf{R} := \begin{bmatrix}
\frac{1}{\sqrt{2}} & - \frac{1}{\sqrt{2}} \\
\frac{1}{\sqrt{2}} & \frac{1}{\sqrt{2}} \\
\end{bmatrix},
\end{equation}
and the diffeomorphism $\diffeoB:\Real^2\to \Real^2$ is given by
\begin{equation}
    \diffeoB(\VectorB):= (\VectorB_1 - h(\VectorB_2), \tanh(\VectorB_2 / 2)),
\end{equation}
where $h:\Real\to\Real$ is given by 
\begin{equation}
    h(x) := \left\{\begin{matrix}
\frac{1}{2} x^2 + \frac{1}{2} & \text{if } |x|\leq 1, \\
|x| & \text{otherwise}. \\
\end{matrix}\right.
\end{equation}

% \subsection{Data-driven Riemannian geometry from a bimodal Gaussian distribution.}
% \label{app:illustrative-tensor-field}

% The anisotropic normal distributions in Figure~\ref{fig:double-gaussian-data-analysis-results-toy-metric} have means 
% $$
%     \mu_1 := (-5, 0)^\top, \quad \mu_2 := (0,5)^\top,
% $$ 
% and covariance matrices 
% $$
%     \spdMatrix_1:= \operatorname{diag}(1/4, 4), \quad \spdMatrix_2 := \operatorname{diag}(4, 1/4).
% $$
% We use these to construct the Riemannian structure
% \begin{equation}
%     (\tangentVector, \tangentVectorB)_{\Vector} := 
%      \sum_{\sumIndA,\sumIndB=1}^2 \frac{e^{-\frac{1}{2}(\Vector - \mu_\sumIndA)^\top \spdMatrix_\sumIndA^{-1}(\Vector - \mu_\sumIndA)-\frac{1}{2}(\Vector - \mu_\sumIndB)^\top\spdMatrix_\sumIndB^{-1}(\Vector - \mu_\sumIndB)}}{\bigl(\sum_{\sumIndC=1}^m e^{-\frac{1}{2}(\Vector - \mu_\sumIndC)^\top \spdMatrix_\sumIndC^{-1}(\Vector - \mu_\sumIndC)}\bigr)^2} (\spdMatrix_\sumIndA^{-1}\tangentVector_\Vector, \spdMatrix_\sumIndB^{-1}\tangentVectorB_\Vector)_2, \quad \Vector \in \Real^2, \tangentVector, \tangentVectorB \in \vectorfield(\Real^2).
%     \label{eq:sum-of-gaussian-Riemannian-geometry}
% \end{equation}
\section{Supplementary material for Section~\ref{sec:iso-riemannian-data-processing}}

\subsection{Proofs of the results in Section~\ref{sec:iso-riemannian-data-processing-mappings}}
\label{app:iso-mapping-dim-1}

\paragraph{Iso-geodesics}
\begin{proof}[Proof of Theorem~\ref{thm:iso-geodesic}]
    We want to show that the mapping 
    \begin{equation}
        t\mapsto \|\frac{\partial}{\partial t} \geodesic^{\iso}_{\Vector,\VectorB}(t)\|_2
    \end{equation}
    is constant.

    First, note that by the chain rule
    \begin{equation}
        \frac{\partial}{\partial t} \geodesic^{\iso}_{\Vector,\VectorB}(t) = \frac{\partial}{\partial t} \geodesic_{\Vector,\VectorB}(\timechange_{\Vector,\VectorB}(t)) = \frac{\partial \timechange_{\Vector,\VectorB}}{\partial t}\dot{\geodesic}_{\Vector,\VectorB}(\timechange_{\Vector,\VectorB}(t)) .
        \label{eq:thm-iso-geodesic-1}
    \end{equation}
    Next, we can evaluate $\frac{\partial \timechange_{\Vector,\VectorB}}{\partial t}$ using the inverse function theorem:
    \begin{equation}
        \frac{\partial \timechange_{\Vector,\VectorB}}{\partial t} = \Bigl( \frac{\partial \timechange_{\Vector,\VectorB}^{-1}}{\partial t'}\Bigr)^{-1}\mid_{t' = \timechange_{\Vector,\VectorB}(t)} = \Bigl( \frac{\|\dot{\geodesic}_{\Vector,\VectorB}(\timechange_{\Vector,\VectorB}(t))\|_2 }{\int_0^1 \|\dot{\geodesic}_{\Vector,\VectorB}(s)\|_2 ds}\Bigr)^{-1} = \frac{\int_0^1 \|\dot{\geodesic}_{\Vector,\VectorB}(s)\|_2 ds}{\|\dot{\geodesic}_{\Vector,\VectorB}(\timechange_{\Vector,\VectorB}(t))\|_2 }.
        \label{eq:thm-iso-geodesic-2}
    \end{equation}
    Combining the above we find that
    \begin{multline}
        \|\frac{\partial}{\partial t} \geodesic^{\iso}_{\Vector,\VectorB}(t)\|_2 \overset{(\ref{eq:thm-iso-geodesic-1})}{=} \|\frac{\partial \timechange_{\Vector,\VectorB}}{\partial t}\dot{\geodesic}_{\Vector,\VectorB}(\timechange_{\Vector,\VectorB}(t))\|_2 \overset{(\ref{eq:thm-iso-geodesic-2})}{=} \| \frac{\int_0^1 \|\dot{\geodesic}_{\Vector,\VectorB}(s) \|_2 ds}{\|\dot{\geodesic}_{\Vector,\VectorB}(\timechange_{\Vector,\VectorB}(t))\|_2 }\dot{\geodesic}_{\Vector,\VectorB}(\timechange_{\Vector,\VectorB}(t))\|_2 \\
        = \int_0^1 \|\dot{\geodesic}_{\Vector,\VectorB}(s)\|_2 ds,
    \end{multline}
    which is constant.
\end{proof}

\paragraph{Iso-logarithms}

\begin{proof}[Proof of Theorem~\ref{thm:iso-log}]
    The first claim (\ref{eq:iso-log-simplified}) follows from direct computation and (\ref{eq:thm-iso-geodesic-2}) from the proof of Theorem~\ref{thm:iso-geodesic}:
    \begin{multline}
     \log_\Vector^{\iso}(\VectorB):= \frac{\partial}{\partial t} \geodesic_{\Vector,\VectorB}(\timechange_{\Vector,\VectorB} (t)) \mid_{t = 0} = \frac{d\tau}{dt}(0) \dot{\geodesic}_{\Vector,\VectorB}(\tau (0)) = \frac{d\tau}{dt}(0) \log_\Vector (\VectorB) \\
     \overset{(\ref{eq:thm-iso-geodesic-2})}{=} \frac{\int_0^1 \|\dot{\geodesic}_{\Vector,\VectorB}(s)\|_2 ds}{\|\log_\Vector (\VectorB)\|_2 }\log_\Vector (\VectorB).
     \label{eq:thm-iso-log}
\end{multline}
The second claim (\ref{eq:iso-log-length}) follows from taking $\ell^2$-norms on both sides of (\ref{eq:thm-iso-log}).
\end{proof}

\paragraph{Iso-exponentials}

\begin{proof}[Proof of Theorem~\ref{thm:iso-exp}]
    To see that $\exp_\Vector^{\iso}(\log_\Vector^{\iso}(\VectorB)) = \VectorB$ holds for $\Vector \neq \VectorB$, it suffices to pick $\vectorchange_{\log_\Vector^{\iso} (\VectorB)} := \frac{\|\log_\Vector (\VectorB)\|_2 }{\int_0^1 \|\dot{\geodesic}_{\Vector,\VectorB}(s)\|_2 ds}$. Indeed, we then have
    \begin{equation}
        \exp_\Vector^{\iso}(\log_\Vector^{\iso}(\VectorB)) \overset{(\ref{eq:iso-exp})}{=} \exp_\Vector(\vectorchange_{\log_\Vector^{\iso} (\VectorB)} \log_\Vector^{\iso}(\VectorB)) \overset{(\ref{eq:iso-log-simplified})}{=} \exp_\Vector(\log_\Vector(\VectorB)) = \VectorB,
        \label{prop:invertibility-iso-log-exp-1}
    \end{equation}
    and 
    \begin{equation}
        \int_{0}^{1}\|\dot{\geodesic}_{\Vector, \exp_\Vector (\vectorchange_{\log_\Vector^{\iso} (\VectorB)} \log_\Vector^{\iso}(\VectorB))}(s)\|_2 \mathrm{d}s \overset{(\ref{prop:invertibility-iso-log-exp-1})}{=} \int_{0}^{1}\|\dot{\geodesic}_{\Vector, \VectorB}(s)\|_2 \mathrm{d}s \overset{(\ref{eq:iso-log-length})}{=} \|\log_\Vector^{\iso}(\VectorB)\|_2.
    \end{equation}
    The case for $\Vector = \VectorB$ follows directly.
    
    Next, we see that $\log_\Vector^{\iso}(\exp_\Vector^{\iso}(\tangentVector_\Vector)) = \tangentVector_\Vector$ for $\tangentVector_\Vector \neq \mathbf{0}$ by direct computation 
    \begin{multline}
        \log_\Vector^{\iso}(\exp_\Vector^{\iso}(\tangentVector_\Vector)) \overset{(\ref{eq:iso-log-simplified})}{=} \frac{\int_0^1 \|\dot{\geodesic}_{\Vector,\exp_\Vector^{\iso}(\tangentVector_\Vector)}(s)\|_2 ds}{\|\log_\Vector (\exp_\Vector^{\iso}(\tangentVector_\Vector))\|_2 }\log_\Vector (\exp_\Vector^{\iso}(\tangentVector_\Vector)) \\
        \overset{(\ref{eq:iso-exp})}{=} \frac{\int_0^1 \|\dot{\geodesic}_{\Vector,\exp_\Vector (\vectorchange_{\tangentVector_\Vector} \tangentVector_\Vector)}(s)\|_2 ds}{\|\log_\Vector (\exp_\Vector (\vectorchange_{\tangentVector_\Vector} \tangentVector_\Vector))\|_2 }\log_\Vector (\exp_\Vector (\vectorchange_{\tangentVector_\Vector} \tangentVector_\Vector)) 
        \overset{(\ref{eq:iso-exp-criterion})}{=} \frac{\|\tangentVector_\Vector\|_{2}}{\|\vectorchange_{\tangentVector_\Vector}\tangentVector_\Vector\|_{2}} \vectorchange_{\tangentVector_\Vector}\tangentVector_\Vector = \tangentVector_\Vector.
    \end{multline}
    The case for $\tangentVector_\Vector = \mathbf{0}$ follows directly.
\end{proof}

\paragraph{Iso-distances}
\begin{proof}[Proof of Theorem~\ref{thm:iso-distance}]
    Both claims follow from direct computation:
    \begin{equation}
        \|\log_\Vector^{\iso}(\VectorB)\|_2 \overset{(\ref{eq:iso-log-simplified})}{=} \int_0^1 \|\dot{\geodesic}_{\Vector,\VectorB}(s)\|_2 ds \overset{(\ref{eq:iso-distance})}{=} \distance_{\Real^\dimInd}^{\iso}(\Vector, \VectorB),
    \end{equation}
    and
    \begin{equation}
        \distance_{\Real^\dimInd}^{\iso}(\Vector, \exp_\Vector^{\iso}(\tangentVector_\Vector)) \overset{(\ref{eq:thm-iso-distance-log})}{=} \|\log_\Vector^{\iso}(\exp_\Vector^{\iso}(\tangentVector_\Vector))\|_2 \overset{(\ref{eq:thm-iso-exp-inv})}{=} \|\tangentVector_\Vector\|_{2}.
    \end{equation}
\end{proof}

\paragraph{Iso-parallel transport}
\begin{proof}[Proof of Theorem~\ref{thm:iso-pt}]
    The first claim (\ref{eq:iso-pt-simplified}) follows directly from subsituting (\ref{eq:thm-iso-geodesic-2}) from the proof of Theorem~\ref{thm:iso-geodesic}.

    For the second claim (\ref{eq:iso-pt-length-geo}), first note that
    \begin{equation}
        \|\mathcal{P}^{\iso}_{\VectorB\leftarrow \Vector}(\dot{\geodesic}^{\iso}_{\Vector,\VectorB}(0))\|_2 \overset{(\ref{eq:iso-pt-length-geo})}{=} \|\dot{\geodesic}^{\iso}_{\Vector,\VectorB}(1)\|_2  \overset{\text{Theorem~\ref{thm:iso-geodesic}}}{=} \|\dot{\geodesic}^{\iso}_{\Vector,\VectorB}(0)\|_2.
    \end{equation}
\end{proof}

% \begin{proof}[Proof of Theorem~\ref{thm:iso-cov}]
%     The claim (\ref{eq:iso-pt-simplified}) follows directly from subsituting (\ref{eq:thm-iso-geodesic-2}) from the proof of Theorem~\ref{thm:iso-geodesic}.
% \end{proof}

\begin{proof}[Proof of Theorem~\ref{thm:iso-cov-geo}]
    The result follows from direct computation:
    \begin{multline}
        \nabla^{\iso}_{\dot{\geodesic}_{\Vector,\VectorB}^{\iso}(t)} \dot{\geodesic}_{\Vector,\VectorB}^{\iso} \\
        = \frac{1}{\frac{\partial \timechange_{\Vector,\VectorB}}{\partial t} (\timechange_{\Vector,\VectorB}(t))\|\dot{\geodesic}_{\Vector,\VectorB}(\timechange_{\Vector,\VectorB}(t))\|_2} \nabla_{\frac{\partial \timechange_{\Vector,\VectorB}}{\partial t} (\timechange_{\Vector,\VectorB}(t)) \dot{\geodesic}_{\Vector,\VectorB}(\timechange_{\Vector,\VectorB}(t))} \Bigl( \|\mathcal{P}_{(\cdot) \leftarrow \geodesic_{\Vector,\VectorB}(\timechange_{\Vector,\VectorB}(t))} \frac{\partial \timechange_{\Vector,\VectorB}}{\partial t} (\timechange_{\Vector,\VectorB}(t)) \dot{\geodesic}_{\Vector,\VectorB} (\timechange_{\Vector,\VectorB}(t))\|_2 \dot{\geodesic}_{\Vector,\VectorB}^{\iso}\Bigr)\\
        \overset{(\ref{eq:thm-iso-geodesic-2})}{=}
        \frac{\partial \timechange_{\Vector,\VectorB}}{\partial t} (\timechange_{\Vector,\VectorB}(t)) \frac{\nabla_{\dot{\geodesic}_{\Vector,\VectorB}(\timechange_{\Vector,\VectorB}(t))} \Bigl( \|\dot{\geodesic}_{\Vector,\VectorB}\|_2 \frac{\int_0^1 \|\dot{\geodesic}_{\Vector,\VectorB}(s) \|_2 ds}{\|\dot{\geodesic}_{\Vector,\VectorB}\|_2 }\dot{\geodesic}_{\Vector,\VectorB}\Bigr)}{\|\dot{\geodesic}_{\Vector,\VectorB}(\timechange_{\Vector,\VectorB}(t))\|_2} \\
        = \frac{\partial \timechange_{\Vector,\VectorB}}{\partial t} (\timechange_{\Vector,\VectorB}(t)) \frac{ \int_0^1 \|\dot{\geodesic}_{\Vector,\VectorB}(s) \|_2 ds}{\|\dot{\geodesic}_{\Vector,\VectorB}(\timechange_{\Vector,\VectorB}(t))\|_2} \nabla_{\dot{\geodesic}_{\Vector,\VectorB}(\timechange_{\Vector,\VectorB}(t))}  \dot{\geodesic}_{\Vector,\VectorB} \\
        = \frac{\partial \timechange_{\Vector,\VectorB}}{\partial t} (\timechange_{\Vector,\VectorB}(t)) \frac{ \int_0^1 \|\dot{\geodesic}_{\Vector,\VectorB}(s) \|_2 ds}{\|\dot{\geodesic}_{\Vector,\VectorB}(\timechange_{\Vector,\VectorB}(t))\|_2} \mathbf{0} = \mathbf{0}.
    \end{multline}
\end{proof}

\begin{proof}[Proof of Theorem~\ref{thm:iso-cov-pt}]
    The result follows from direct computation:
    \begin{multline}
        \nabla^{\iso}_{\dot{\geodesic}_{\Vector,\VectorB}(t)} \tangentVector = \frac{\nabla_{\dot{\geodesic}_{\Vector,\VectorB}(t)} \Bigl( \|\dot{\geodesic}_{\Vector,\VectorB}\|_2 \frac{\|\dot{\geodesic}_{\Vector,\VectorB}(0)\|_2}{\|\dot{\geodesic}_{\Vector,\VectorB}\|_2 }\mathcal{P}_{(\cdot)\leftarrow \Vector}(\tangentVector_\Vector)\Bigr)}{\|\dot{\geodesic}_{\Vector,\VectorB}(t)\|_2}  
        = \frac{\|\dot{\geodesic}_{\Vector,\VectorB}(0)\|_2}{\|\dot{\geodesic}_{\Vector,\VectorB}(t)\|_2} \nabla_{\dot{\geodesic}_{\Vector,\VectorB}(t)}  \mathcal{P}_{(\cdot)\leftarrow \Vector}(\tangentVector_\Vector) \\
        =  \frac{\|\dot{\geodesic}_{\Vector,\VectorB}(0)\|_2}{\|\dot{\geodesic}_{\Vector,\VectorB}(t)\|_2}  \mathbf{0} = \mathbf{0}.
    \end{multline}
\end{proof}

\subsection{Numerical approximations of the mappings in Section~\ref{sec:iso-riemannian-data-processing-mappings}}
\label{app:iso-mapping-dim-2}
Assuming that we have all manifold mappings in closed-form (which is the case for Euclidean pullback metrics), it remains to construct numerical approximations for iso-geodesics, iso-logarithms and iso-exponential mappings. The iso-distance can be found through the iso-logarithm and the iso-parallel transport mapping can be computed directly.

\paragraph{Iso-geodesics}

In general it will be hard to retrieve $\timechange_{\Vector,\VectorB}$ from its inverse 
\begin{equation}
    \timechange_{\Vector,\VectorB}^{-1}(t'):= \frac{\int_0^{t'} \|\dot{\geodesic}_{\Vector,\VectorB}(s)\|_2 ds}{\int_0^1 \|\dot{\geodesic}_{\Vector,\VectorB}(s)\|_2 ds}.
\end{equation}

To get an approximate inverse, we first approximate the geodesic as a discrete curve consisting of $M+1\in \Natural$ points (including end points), i.e., we construct the curve $\geodesic_{\Vector,\VectorB}^M:[0,1]\to \Real^\dimInd$ given by
\begin{equation}
    \geodesic_{\Vector,\VectorB}^M(t) := (1 - M(t - \lfloor tM\rfloor / M))\geodesic_{\Vector,\VectorB}(\frac{\lfloor tM\rfloor}{M}) + M(t - \lfloor tM\rfloor / M) \geodesic_{\Vector,\VectorB}(\frac{\lceil tM\rceil}{M}).
\end{equation}

For such a discrete geodesic, we can find an expression for $\timechange_{\Vector,\VectorB}^M$ in closed-form:
\begin{equation}
    \timechange_{\Vector,\VectorB}^M(t) := \frac{t \sum_{\sumIndA=1}^{M} \|\geodesic_{\Vector,\VectorB}(\frac{\sumIndA}{M}) - \geodesic_{\Vector,\VectorB}(\frac{\sumIndA-1}{M})\|_2 - \sum_{\sumIndA=1}^{K_t} \|\geodesic_{\Vector,\VectorB}(\frac{\sumIndA}{M}) - \geodesic_{\Vector,\VectorB}(\frac{\sumIndA-1}{M})\|_2}{\|\geodesic_{\Vector,\VectorB}(\frac{K_t+1}{M}) - \geodesic_{\Vector,\VectorB}(\frac{K_t}{M})\|_2} + \frac{K_t}{M},
\end{equation}
where 
\begin{equation}
    K_t := \inf \{K' \in \Natural \mid t \sum_{\sumIndA=1}^{M} \|\geodesic_{\Vector,\VectorB}(\frac{\sumIndA}{M}) - \geodesic_{\Vector,\VectorB}(\frac{\sumIndA-1}{M})\|_2 \geq \sum_{\sumIndA=1}^{K} \|\geodesic_{\Vector,\VectorB}(\frac{\sumIndA}{M}) - \geodesic_{\Vector,\VectorB}(\frac{\sumIndA-1}{M})\|_2\}.
\end{equation}

For $M$ large enough, we have $\timechange_{\Vector,\VectorB} \approx \timechange_{\Vector,\VectorB}^M$ and $\geodesic^{\iso}_{\Vector,\VectorB}(t) \approx \geodesic_{\Vector,\VectorB}(\timechange^M_{\Vector,\VectorB}(t))$, which is the approximation we use. The default value $M=100$ is used in this work.

\paragraph{Iso-logarithms}
For iso-logarithms we use that 
\begin{equation}
    \int_0^1 \|\dot{\geodesic}_{\Vector,\VectorB}(s)\|_2 ds \approx \sum_{\sumIndA=1}^{M} \|\geodesic_{\Vector,\VectorB}(\frac{\sumIndA}{M}) - \geodesic_{\Vector,\VectorB}(\frac{\sumIndA-1}{M})\|_2
\end{equation}
for large $M\in \Natural$, which gives the approximation
\begin{equation}
    \log_\Vector^{\iso}(\VectorB) \approx \frac{\sum_{\sumIndA=1}^{M} \|\geodesic_{\Vector,\VectorB}(\frac{\sumIndA}{M}) - \geodesic_{\Vector,\VectorB}(\frac{\sumIndA-1}{M})\|_2}{\|\log_\Vector (\VectorB)\|_2 }\log_\Vector (\VectorB).
\end{equation}
Again the default value $M=100$ is used in this work.

\paragraph{Iso-exponentials}
Finally, for iso-exponential mappings we use the approximation
\begin{equation}
    \exp_\Vector^{\iso}(\tangentVector_\Vector) \approx \Bigl(1 - \frac{\|\tangentVector_\Vector\|_2 - \sum_{\sumIndA=1}^{K-1} \|\VectorC^\sumIndA - \VectorC^{\sumIndA-1}\|_2}{\|\VectorC^K - \VectorC^{K-1}\|_2}\Bigr)\VectorC^K + \frac{\|\tangentVector_\Vector\|_2 - \sum_{\sumIndA=1}^{K-1} \|\VectorC^\sumIndA - \VectorC^{\sumIndA-1}\|_2}{\|\VectorC^K - \VectorC^{K-1}\|_2}\VectorC^{K-1},
\end{equation}
where $\VectorC^0 := \Vector$, $\VectorC^1 := \exp_{\Vector}(\frac{1}{M} \tangentVector_\Vector)$ (here too $M=100$ is used) and
\begin{equation}
    \VectorC^\sumIndA := \geodesic_{\VectorC^{\sumIndA-2},\VectorC^{\sumIndA-1}} (2), \quad \sumIndA = 2, \ldots K,
\end{equation}
where 
\begin{equation}
    K := \inf \{K'\in \Natural \mid  \sum_{\sumIndA=1}^{K'} \|\VectorC^\sumIndA - \VectorC^{\sumIndA-1}\|_2 - \|\tangentVector_\Vector\|_2 \geq 0\}.
\end{equation}

\subsection{Proof of the result in Section~\ref{sec:iso-riemannian-data-processing-low-rank}}
\label{app:iso-mapping-dim-3}

\begin{proof}[Proof of Theorem~\ref{thm:global-approximation-error}]
    First, we define the geodesic variation $\Gamma:[0,1]\times[-\epsilon, \epsilon] \to \Real^{\dimInd \times \dataPointNum}$  as
\begin{equation}
    \Gamma(t,s):= \exp_{\mPoint}\Biggl( \isoDiffeo_\mPoint \biggl(t \Bigl(\isoDiffeo_\mPoint^{-1} (\log_\mPoint(\Matrix)) + \frac{s}{\epsilon}\bigr(\tangentVector_\mPoint - \isoDiffeo_\mPoint^{-1} (\log_\mPoint(\Matrix))\bigl)\Bigr)\biggr)\Biggr), 
\end{equation}
where $\epsilon := \| \isoDiffeo_\mPoint^{-1} (\log_\mPoint(\Matrix)) - \tangentVector_\mPoint\|_{F}$.
Note that $\Gamma(1,0) = \Matrix$ and $\Gamma(1,\epsilon) = \exp_{\mPoint}(\isoDiffeo_\mPoint(\tangentVector_\mPoint))$.

We will show that (\ref{eq:jacobi-bound-low-rank}) holds through Taylor expansion of $\|\Gamma(1,0), \Gamma(1,s)\|_F^2$ around $s=0$. Expanding up to second order gives
\begin{multline}
    \|\Matrix - \exp_{\mPoint}(\isoDiffeo_\mPoint (\tangentVector_\mPoint))\|_F^2 = \|\Gamma(1,0)- \Gamma(1,\epsilon)\|_F^2 
    = \|\Gamma(1,0)- \Gamma(1,0)\|_F^2 \\
    +  \epsilon \frac{\mathrm{d}}{\mathrm{d} s} \|\Gamma(1,0),-\Gamma(1,s)\|^2\mid_{s=0} 
    + \frac{\epsilon^2}{2} \frac{\mathrm{d}^2}{\mathrm{d} s^2} \|\Gamma(1,0)- \Gamma(1,s)\|_F^2\mid_{s=0} \\
    + \frac{1}{2}\int_0^\epsilon \frac{\mathrm{d}^3}{\mathrm{d} s'^3} \|\Gamma(1,0)- \Gamma(1,s')\|_F^2 (s - s')^2\mathrm{d}s'.
    \label{eq:jacobi-bound-low-rank-thm-taylor}
\end{multline}
We immediately see that 
\begin{equation}
    \|\Gamma(1,0)- \Gamma(1,0)\|_F^2=0.
    \label{eq:jacobi-bound-low-rank-thm-taylor0}
\end{equation}
So for proving the result (\ref{eq:jacobi-bound-low-rank}), we must show that 
\begin{equation}
    \frac{\mathrm{d}}{\mathrm{d} s} \|\Gamma(1,0)- \Gamma(1,s)\|_F^2\mid_{s=0} = 0,
    \label{eq:jacobi-bound-low-rank-thm-taylor1}
\end{equation}
and that the second-order term in (\ref{eq:jacobi-bound-low-rank-thm-taylor}) reduces to the first term in (\ref{eq:jacobi-bound-low-rank}).
It is clear that the remainder term is $\mathcal{O}(\epsilon^3)$.

Notice that (\ref{eq:jacobi-bound-low-rank-thm-taylor1}) follows from
\begin{equation}
    \frac{\mathrm{d}}{\mathrm{d} s} \|\Gamma(1,0)- \Gamma(1,s)\|_F^2 = 2(\frac{\partial}{\partial s} \Gamma(1,s), \Gamma(1,s) - \Gamma(1,0))_F
    \label{eq:towards-jacobi-bound-low-rank-thm-taylor1}
\end{equation}
by evaluating \cref{eq:towards-jacobi-bound-low-rank-thm-taylor1} at $s=0$. Next, from \cref{eq:towards-jacobi-bound-low-rank-thm-taylor1} we find that
\begin{equation}
    \frac{\mathrm{d}^2}{\mathrm{d} s^2} \|\Gamma(1,0)- \Gamma(1,s)\|_F^2 = 2(\frac{\partial^2}{\partial s^2} \Gamma(1,s), \Gamma(1,s) - \Gamma(1,0))_F + 2(\frac{\partial}{\partial s} \Gamma(1,s), \frac{\partial}{\partial s}\Gamma(1,s))_F.
    \label{eq:towards-jacobi-bound-low-rank-mfld-tensor-thm-taylor2}
\end{equation}

Evaluating \cref{eq:towards-jacobi-bound-low-rank-mfld-tensor-thm-taylor2} at $s=0$ gives
\begin{multline}
    \frac{\mathrm{d}^2}{\mathrm{d} s^2} \|\Gamma(1,0)- \Gamma(1,s)\|_F^2 \mid_{s=0} = 2\|\frac{\partial}{\partial s} \Gamma(1,0)\|_F^2 \\
    = 2 \|D_{\isoDiffeo_\mPoint^{-1} (\log_\mPoint(\Matrix))} \exp_{\mPoint}(\isoDiffeo_\mPoint (\cdot)) [\frac{1}{\epsilon}\bigr(\tangentVector_\mPoint - \isoDiffeo_\mPoint^{-1} (\log_\mPoint(\Matrix))\bigl)]\|_F^2\\
    = \frac{2}{\epsilon^2} \sum_{\sumIndA=1}^\dataPointNum \|D_{\isoDiffeo_\mPoint^{-1} (\log_\mPoint(\Vector^\sumIndA))} \exp_{\mPoint}(\isoDiffeo_\mPoint (\cdot)) [\frac{1}{\epsilon}\bigr(\tangentVector_\mPoint - \isoDiffeo_\mPoint^{-1} (\log_\mPoint(\Vector^\sumIndA))\bigl)]\|_2^2\\
    = \frac{2}{\epsilon^2} \sum_{\sumIndA=1}^\dataPointNum  (\isoDiffeo_\mPoint^{-1}(\log_{\mPoint}\Vector^\sumIndA) - \tangentVectorComp^\sumIndA_\mPoint)^\top\mathbf{M}_{\isoDiffeo_\mPoint}^\sumIndA(\isoDiffeo_\mPoint^{-1}(\log_{\mPoint}\Vector^\sumIndA) - \tangentVectorComp^\sumIndA_\mPoint).
\end{multline}
When the above is substituted back into (\ref{eq:jacobi-bound-low-rank-thm-taylor}), the second-order term reduces to the first term in (\ref{eq:jacobi-bound-low-rank}), which proves the claim.
    
\end{proof}
\section{Training details}
\label{app:training-details}

\paragraph{Common parameters}
\begin{itemize}
    \item \textbf{Optimizer}: Adam with \texttt{betas} = (0.9, 0.99) and learning rate $10^{-3}$.
\end{itemize}

\paragraph{Architecture parameters}
\begin{itemize}
    \item Double Gaussian:
    \begin{itemize}
        \item $\diffeoD_{\networkParams_2^\sumIndC}$: 
        \begin{itemize}
            \item Number of Householder layers: 2
        \end{itemize}
        \item $f_{\networkParams_1^\sumIndC}$: 
        \begin{itemize}
            \item Masked 1D convolution: fixed filter $[1,0,1]$
            \item Activation: order = 2
        \end{itemize} 
    \end{itemize}
    \item Hemisphere:
    \begin{itemize}
        \item $\diffeoD_{\networkParams_2^\sumIndC}$: 
        \begin{itemize}
            \item Number of Householder layers: 3
        \end{itemize}
        \item $f_{\networkParams_1^\sumIndC}$: 
        \begin{itemize}
            \item Masked 1D convolution: fixed filter $[1,0,1]$
            \item Activation: order = 2
        \end{itemize} 
    \end{itemize}
    \item MNIST:
    \begin{itemize}
        \item $\diffeoD_{\networkParams_2^\sumIndC}$: 
        \begin{itemize}
            \item Masked 2D convolutions: input channels = 1, output channel = 1, kernel size = 5 and padding = 2
        \end{itemize}
        \item $f_{\networkParams_1^\sumIndC}$: 
        \begin{itemize}
            \item Masked 2D convolution: input channels = 1, output channel = 128, kernel size = 5, padding = 2
            \item Activation: order = 6
            \item Masked 2D convolution: input channels = 128, output channel = 128, kernel size = 5, padding = 2
            \item Activation: order = 6
            \item Masked 2D convolution: input channels = 128, output channel = 1, kernel size = 5, padding = 2
        \end{itemize}
    \end{itemize}
\end{itemize}

\paragraph{Remaining parameters}
The remaining parameters are summarized below:
\begin{table}[h!]
    \centering
    \begin{tabular}{l|l|l|l|l}
         \hline
\textbf{Data set} & \textbf{Flow steps} &  \textbf{Epochs} & \textbf{Batch size} & \textbf{Weight decay}  \\ \hline
Double Gaussian & 2 & 500 & 16 & 0.2 \\ \hline
Hemisphere & 3 & 500 & 16 & 0.02 \\ \hline
MNIST & 6 & 100 & 128 & 0. \\ \hline \hline
    \end{tabular}
\end{table}

\end{document}